%% file: neurips_2026.tex
\documentclass{article}

\usepackage[main,final]{neurips_2026}

\usepackage[utf8]{inputenc} 
\usepackage[T1]{fontenc}    
\usepackage{hyperref}       
\usepackage{url}            
\usepackage{booktabs}       
\usepackage{amsfonts}       
\usepackage{nicefrac}       
\usepackage{microtype}      
\usepackage[table]{xcolor}  
\usepackage{comment}
\usepackage{wrapfig}
\usepackage{caption}
  \usepackage{amsmath,amssymb,amsthm}
  \usepackage{bm}
  \usepackage{algorithm}
  \usepackage{algpseudocode}
\usepackage{graphicx}
\usepackage{enumitem}
\usepackage{multirow}
\usepackage{xspace}
\newcommand{\pending}{\textcolor{gray}{\textit{pend.}}}
\newcommand{\RESULTS}[1]{\pending}

\newcommand{\CGRPO}{\texorpdfstring{\textup{\textsc{C-GRPO}}\xspace}{C-GRPO}}
\newcommand{\CGRPOloss}{\mathcal{L}_{\textrm{C-GRPO}}}
  \newtheorem{theorem}{Theorem}
  \newtheorem{lemma}[theorem]{Lemma}
  \newtheorem{corollary}[theorem]{Corollary}
  \newtheorem{proposition}[theorem]{Proposition}
  \newtheorem{definition}[theorem]{Definition}
  \newtheorem{remark}[theorem]{Remark}

\hypersetup{
  hypertexnames=false,
  colorlinks=true,
  citecolor=blue,
  linkcolor=red,
  urlcolor=blue
}

\title{C-GRPO: Conformal Group Relative Policy Optimization}

\author{%
  Arya Fayyazi\thanks{Corresponding author: \texttt{afayyazi@usc.edu}} \And
  Seyedarmin Azizi \And
  Mehdi Kamal \And
  Massoud Pedram \AND
  University of Southern California \\
  Los Angeles, CA, USA
}

\begin{document}

\maketitle

\input{sec/0_abstract}    
\input{sec/1_intro}
\input{sec/2_background}
\input{sec/3_method}
\input{sec/4_results}
\input{sec/5_conclusion}

\section*{Acknowledgements and Disclosures}
\textbf{Funding.} This work received no external funding or third-party support during the last 36 months.
\textbf{Competing interests.} The authors declare no competing interests: they have no financial relationships outside the submitted work that could be perceived as influencing it.

\bibliography{main}
\bibliographystyle{plainnat}
\newpage
\appendix
\input{sec/A_Appendix}



\newpage
\input{checklist.tex}

\end{document}

%% file: sec/0_abstract.tex
\begin{abstract}
Reasoning-oriented language model training increasingly relies on multiple on-policy rollouts, but standard Group Relative Policy Optimization (GRPO) uses a fixed group size, wasting compute on easy prompts and under-exploring hard ones. Existing adaptive-budget methods alleviate this inefficiency, yet their stopping rules are heuristic and do not provide a finite-sample calibration guarantee. We propose \CGRPO, which replaces the fixed rollout budget of standard GRPO with calibrated conformal thresholds over task-specific nonconformity scores, combining Adaptive Prediction Set (APS)-style calibration for closed-form reasoning, a first-success score for execution-based code, automatic miscoverage selection, and periodic re-calibration. We prove finite-sample false-stop control at every budget and simultaneous coverage for the adaptively selected budget under multiplicity-corrected calibration, and show that, under monotone policy improvement, recalibration induces a non-increasing expected training budget (an idealized tendency). Across six benchmarks spanning mathematics, graduate-level science, and code generation, \CGRPO reduces mean training rollouts by 49--70\% relative to fixed-budget GRPO at matched accuracy, with pass@1 equal to or slightly above that of the evaluated static and dynamic baselines (differences within binomial noise), and transfers across model families with up to 82\% rollout savings.
\end{abstract}

%% file: sec/1_intro.tex
\vspace{-1em}
\section{Introduction}
\label{sec:intro}
\vspace{-1em}

As LLMs scale, a massive shift toward multi-sample computation at both inference and training time is taking place. Strategies like majority voting for reasoning tasks and best-of-$k$ sampling for code generation consistently push the Pareto frontier. Even modern RL algorithms such as GRPO (Group Relative Policy Optimization)~\citep{shao2024deepseekmath} leverage this by using on-policy rollouts (generating a batch of responses for the same prompt) to derive a relative advantage signal for the loss function. However, the bottleneck is that GRPO currently uses a static group size ($G$) for every input. This is fundamentally inefficient because it ignores the high variance in task difficulty. For ``low-entropy" or trivial queries, generating a large group is just wasted FLOPs with a near-zero gradient signal. In contrast, for ``frontier" or high-complexity queries, a fixed $G$ is often too small to capture a single successful trajectory, leaving the model with no positive reinforcement to learn from.

Adaptive-budget methods address this with reward variance, learned difficulty estimates, or global allocation rules~\citep{zhang2026aero, panaganti2026gdro, xiong2025reinforceada, zheng2025greso}, but their stopping rules remain heuristic and offer no finite-sample guarantee as the policy evolves. Conformal prediction provides distribution-free uncertainty certificates \citep{vovk2005algorithmic, romano2020classification}, but has been used only as a post-hoc inference wrapper, not as a training-time compute control mechanism~\citep{quach2024conformal, gui2024alignment, cherian2024validity}.

We bridge these perspectives with \textbf{Conformal GRPO} (\CGRPO), which replaces GRPO's fixed rollout budget with calibrated stopping thresholds.  For each budget $k \in \mathcal{K}$, \CGRPO selects the smallest $k$ which is statistically sufficient for the current prompt. 
Practical ingredients include Adaptive Prediction Set (APS) calibration for closed-form reasoning, a new \emph{first-success} score for execution-based code,  automatic miscoverage selection to prevent degenerate early-training thresholds, 
and periodic recalibration to track the evolving policy. 
The result is a self-paced curriculum: easy queries exit early, and computational resources are reallocated to the hard tail.

We prove finite-sample coverage for the calibrated thresholds, derive false-stop control at every budget and a multiplicity-corrected guarantee for the adaptively selected budget, and establish a non-increasing expected training budget under monotone policy improvement (an idealized condition; Sec.~\ref{sec:efficiency}). Empirically, across six benchmarks that span mathematics, graduate-level science, and code, \CGRPO reduces mean training rollouts by \textbf{49--70\%} compared to fixed-budget GRPO at matched accuracy: pass@1 on GSM8K, HumanEval, MBPP, and MATH-500 is equal to or slightly above every evaluated baseline (differences are within binomial noise) with the fewest rollouts per prompt; rollout savings further reach \textbf{82\%} on GSM8K for Llama-3.1-8B without re-tuning. These theoretical and empirical results are enabled by three contributions: \textbf{(i)} we formulate adaptive-group GRPO as a conformal calibration problem; \textbf{(ii)} we introduce practical ingredients for LLM RL, including APS with first-success scores, auto-$\delta$, and periodic re-calibration; and \textbf{(iii)} we prove coverage, false-stop and self-annealing guarantees and show empirically that calibrated allocation matches the accuracy of fixed and heuristic budgets at far fewer rollouts, including fixed-$G$ GRPO at matched compute.

%% file: sec/2_background.tex
\vspace{-1em}
\section{Background}
\label{sec:background}
\vspace{-0.5em}
\subsection{Policy Optimization and GRPO}
\label{sec:bg_grpo}

Group Relative Policy Optimization (GRPO) \citep{shao2024deepseekmath} eliminates the memory-heavy value network of the PPO (Proximal Policy Optimization) \citep{schulman2017proximal} by computing advantages from a fixed group of $G$ i.i.d.\ (independent and identically distributed) completions $\{y_1,\ldots,y_G\}\sim\pi_\theta(\cdot\mid x)$:
\begin{equation}
    \hat{A}^{\text{GRPO}}_i = \frac{r_i - \bar{r}}{\sigma_r + \epsilon}, \quad \bar{r} = \frac{1}{G}\sum_{j=1}^G r_j, \quad \sigma_r = \sqrt{\frac{1}{G}\sum_{j=1}^G (r_j - \bar{r})^2}.
    \label{eq:bg_grpo_adv}
\end{equation}
where $\epsilon > 0$ is a small numerical stabilizer (e.g., $10^{-8}$).
This achieves state-of-the-art results on mathematical reasoning and code generation.
However, fixing $G$ uniformly wastes compute on easy queries (all $r_j{\approx}1$, zero advantage)
and starves hard queries of exploration (all $r_j{=}0$, no positive signal), ignoring the wide
per-prompt difficulty disparity \citep{damani2025thinking}.
\vspace{-1em}
\subsection{Split-Conformal Prediction}
\label{sec:bg_conformal}

Split-conformal prediction (CP) \citep{vovk2005algorithmic} constructs statistically
guaranteed prediction sets with no distributional assumptions.
Given $n$ i.i.d.\ calibration pairs $(X_i,Y_i)$ and a nonconformity score
$s(X,Y)\in\mathbb{R}$, the empirical $(1{-}\delta)$ quantile $\hat{q}$
(with the standard finite-sample correction $\lceil(n{+}1)(1{-}\delta)\rceil$)
satisfies
$P\bigl(s(X_{n+1},Y_{n+1})\leq\hat{q}\bigr)\in[1{-}\delta,\;1{-}\delta{+}\tfrac{1}{n+1}]$
for any exchangeable test point. Prior work applies CP \emph{post-hoc} for inference-time abstention and safety
filtering \citep{quach2024conformal}.
\CGRPO inverts this paradigm: rather than certifying a fixed model's
outputs, we embed CP inside the training loop so calibrated quantiles
directly control the per-query generation budget, converting uncertainty
measures into online compute decisions.
\vspace{-1em}
\section{Related Work}
\label{sec:related_work}
\vspace{-0.5em}
\paragraph{Efficient RL and Adaptive Rollout Allocation.}
Adaptive rollout methods address GRPO's fixed-$G$ inefficiency via rollout pruning, difficulty-aware allocation, and adaptive sampling \citep{zhang2026aero, panaganti2026gdro, xiong2025reinforceada, zheng2025greso}, but rely on heuristic or model-dependent signals (reward variance, learned difficulty scores) that lack formal guarantees and drift as the policy evolves. \CGRPO replaces them with a distribution-free conformal stopping rule grounded in finite-sample coverage theory.
\vspace{-1em}
\paragraph{Conformal Prediction for LLMs.}
Conformal methods have been applied to LLMs for calibrated uncertainty, abstention, and output filtering \citep{quach2024conformal, gui2024alignment, cherian2024validity}, as static post-hoc wrappers relying on token-level likelihoods that are uninformative when every completion is a unique string. We instead embed CP in the training objective and introduce an ordinal first-success score for execution-based reinforcement-learning (RL) settings.
\vspace{-1em}
\paragraph{Curriculum and Self-Paced Learning.}
Curriculum methods improve convergence by ordering examples from easy to hard \citep{kumar2010self}; adaptive rollout methods implicitly induce such curricula without guarantees.
\CGRPO instantiates a self-paced curriculum (Corollary~\ref{cor:annealing}, under an idealized monotone-improvement condition, and empirically in aggregate): as the policy improves, calibrated thresholds $\hat{q}_k$ tighten, shifting compute to harder queries with no external supervision or schedule.

%% file: sec/3_method.tex
\section{Methodology}
\label{sec:method}

\begin{figure}[!ht]
    \centering
    \includegraphics[width=\linewidth]{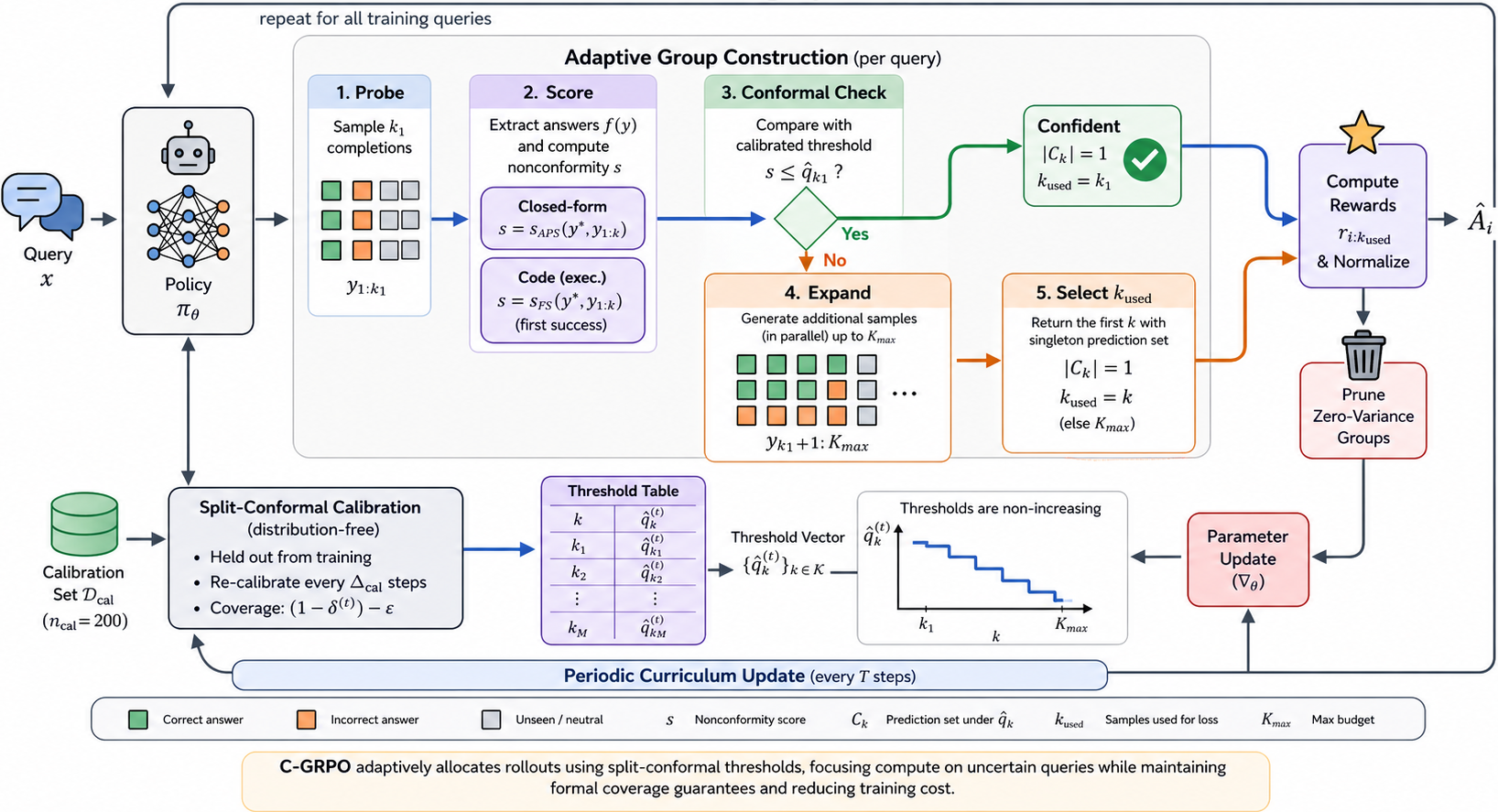}
    \caption{Overview of C-GRPO workflow.}
    \vspace{-1em}
    \label{fig:example}
\end{figure}

\vspace{-0.5em}
\subsection{Problem Formulation}
\label{sec:problem}

Let $\mathcal{X}$ denote the space of natural-language queries and $\mathcal{Y}$
the space of model completions.  A language model defines a conditional
distribution $\pi_\theta(\cdot\mid x)$ over $\mathcal{Y}$ parameterized by
$\theta$.  An answer-extraction function $f:\mathcal{Y}\to\mathcal{A}$ maps
raw completions to a structured answer space $\mathcal{A}$, and a deterministic
verifier $v:\mathcal{A}\times\mathcal{A}\to\{0,1\}$ checks correctness against
the gold answer $y^{\star}$:
\begin{equation}
  v(a,\, a^{\star}) =
  \begin{cases}
    1 & \text{if } a \text{ is verified correct w.r.t.\ } a^{\star},\\
    0 & \text{otherwise.}
  \end{cases}
\end{equation}
We consider two task families: \textbf{closed-form} (math, science), where $f$
extracts a canonicalized answer and $v$ is exact-match; and \textbf{generative
code} (HumanEval, MBPP), where $v$ executes against hidden unit tests.
Let $\mathcal{K}{=}\{k_1{<}\cdots{<}k_M\}$ be a fixed budget grid with
$K_{\max}{=}k_M$ and rewards $r_j{=}v(f(\hat{y}_j),y^{\star})$.
The training objective is:
\begin{equation}
  \theta^{\star}
    = \arg\max_\theta \;
      \mathbb{E}_{x}\!\left[\mathbb{E}\!\left[\max_{j\leq k_{\mathrm{used}}} r_j\right]\right]
    \quad \text{s.t.} \quad
      \mathbb{E}_{x}\bigl[k_{\mathrm{used}}(x;\theta)\bigr] \leq C.
  \label{eq:objective}
\end{equation}
Here $C > 0$ is a target per-step compute budget, and $k_{\mathrm{used}}(x;\theta)$ is the adaptive group size for query $x$ under policy $\pi_\theta$ (formally defined in Section~\ref{sec:adaptive_group}).
\CGRPO solves this without explicit constraint optimization: calibrated
conformal thresholds implicitly determine $k_{\mathrm{used}}$ per query.
\vspace{-1em}
\subsection{Conformal Nonconformity Scores}
\label{sec:scores}

The conformal machinery requires a \emph{nonconformity score}
$s:\mathcal{A}\times\mathcal{A}^k\to[0,1]$ quantifying how surprising the gold
answer $y^{\star}$ is given $k$ model samples.  A good score must be
(i)~\emph{continuous} so that calibrated quantiles strictly decrease as the
model improves, and (ii)~\emph{consistent with exchangeability} of the samples
so that split-conformal validity holds without distributional assumptions on $P$.
\vspace{-1em}
\subsubsection{APS Score for Closed-Form Answers}
\label{sec:aps_score}

For closed-form tasks we use the \textbf{Adaptive Prediction Sets} (APS) score
of \citet{romano2020classification}.  Given $k$ extracted answers
$\hat{a}_1,\ldots,\hat{a}_k$, let $\hat{p}(a) = k^{-1}\sum_{j=1}^k
\mathbf{1}[\hat{a}_j = a]$ be the empirical frequency.  Sort distinct answers
in decreasing frequency order $a_{\sigma(1)}, a_{\sigma(2)},\ldots$
(ties broken uniformly at random to strictly preserve statistical exchangeability).
The APS nonconformity score is
\begin{equation}
  s_{\mathrm{APS}}(y^{\star};\,\hat{a}_{1:k})
    \;=\;
    \underbrace{\vphantom{\Big[}
      \sum_{j:\,\hat{p}(a_{\sigma(j)}) > \hat{p}(y^{\star})}
        \hat{p}(a_{\sigma(j)})
    }_{\text{mass strictly more frequent than gold}}
    \;+\;
    \underbrace{\vphantom{\Big[}
      U \cdot \hat{p}(y^{\star})
    }_{\text{randomized fractional mass}},
    \quad U \sim \mathrm{Uniform}[0,1],
  \label{eq:aps_score}
\end{equation}
with $s_{\mathrm{APS}} = 1$ when $\hat{p}(y^{\star}) = 0$ (gold never sampled).

The APS score is strictly superior to the naive binary score
$s_{\mathrm{bin}} = 1 - \hat{p}(y^{\star})$: the $U$-randomization makes it
\emph{strictly continuous} on $[0,1]$, ensuring the $(1-\delta)$ quantile
remains informative even for weak models, and achieves
\emph{exact} (not merely conservative) marginal coverage \citep{romano2020classification}.
\vspace{-1em}
\subsubsection{First-Success Score for Code}
\label{sec:code_score}

For execution-based code tasks, every generated completion is a syntactically
unique string, so $\hat{p}(y^{\star}) = 0$ almost surely and
$s_{\mathrm{APS}} = 1$ for all calibration examples regardless of budget,
making the APS framework vacuous.  We therefore introduce a novel
\textbf{first-success score} based on the \emph{ordinal position} of the first
passing solution within the ordered sample sequence:
\begin{equation}
  s_{\mathrm{code}}(y^{\star};\,\hat{y}_{1:k})
    \;=\;
    \begin{cases}
      \dfrac{j^{\star}}{k},
        & j^{\star} = \min\!\bigl\{j : v(\hat{y}_j, y^{\star})=1\bigr\}
          \;\text{exists,} \\[8pt]
      1,
        & \text{no sample passes all unit tests.}
    \end{cases}
  \label{eq:code_score}
\end{equation}
That is, the score is the fraction of the budget consumed before the first
success.  It is continuous on $(0,1]$: an easy problem solved on the first
of $k$ samples scores $1/k \to 0$; an unsolvable problem always scores $1$.

This score has a key calibration property: $\hat{q}_k^{\mathrm{code}} < 1$ if
and only if at least $\lceil(n+1)(1-\delta)\rceil - 1$ calibration problems
have at least one passing solution within $k$ samples (Section~\ref{sec:calibration}).
As the policy improves during training, more problems become solvable within
small $k$, driving $\hat{q}_k$ below 1 and enabling the conformal curriculum
to engage at smaller budgets.
\vspace{-1em}
\paragraph{Exchangeability.}
Both scores are functions of $(y^{\star}, \hat{y}_{1:k})$ and inherit
exchangeability from i.i.d.\ sampling of completions and calibration
examples---the sole assumption conformal validity requires.
\vspace{-1em}
\subsection{Split-Conformal Calibration}
\label{sec:calibration}

Given a held-out calibration set
$\mathcal{D}_{\mathrm{cal}} = \{(x_i,y^{\star}_i)\}_{i=1}^{n}$ disjoint from
training data and a model $\pi$, we generate $K_{\max}$ samples
$\hat{y}_{i,1},\ldots,\hat{y}_{i,K_{\max}}\sim\pi(\cdot\mid x_i)$ per query
and derive one threshold per budget level.

\begin{definition}[Split Conformal Quantile]
  \label{def:conformal_quantile}
  For scores $s_1,\ldots,s_n \in [0,1]$ and miscoverage level $\delta\in(0,1)$,
  the \emph{finite-sample conformal quantile} is:
  \begin{equation}
    \hat{q}(s_{1:n},\,\delta)
      \;=\;
      s_{\bigl(\min\bigl(\lceil(n+1)(1-\delta)\rceil,\,n\bigr)\bigr)},
    \label{eq:conformal_quantile}
  \end{equation}
  where $s_{(j)}$ denotes the $j$-th order statistic.  The correction
  $\lceil(n+1)(1-\delta)\rceil$ (rather than $\lceil n(1-\delta)\rceil$) is
  the standard finite-sample adjustment ensuring
  $P(s_{\mathrm{test}} \leq \hat{q}) \geq 1-\delta$ for an exchangeable test
  score \citep{vovk2005algorithmic}.
\end{definition}

\noindent
For each $k\in\mathcal{K}$ and score $s\in\{s_{\mathrm{APS}},
s_{\mathrm{code}}\}$, the calibrated threshold is:
\begin{equation}
  \hat{q}_k
    \;=\;
    \hat{q}\!\left(
      \bigl\{s(y^{\star}_i;\,\hat{y}_{i,1:k})\bigr\}_{i=1}^{n},
      \;\delta
    \right).
  \label{eq:qhat}
\end{equation}
The result is a threshold vector $\bm{\hat{q}} = (\hat{q}_{k_1},\ldots,\hat{q}_{k_M})$
encoding the model's difficulty profile across budgets.

\paragraph{Properties.}
Monotonicity: $\hat{q}_{k_1}\geq\hat{q}_{k_2}$ for $k_1{\leq}k_2$, so
$\bm{\hat{q}}$ forms a decreasing staircase marking where the policy transitions
from uncertain to confident.
Sensitivity: as $\pi_\theta$ improves, $\hat{q}_k$ decreases for small $k$,
making calibrated thresholds a \emph{live signal of policy improvement} that
justifies periodic re-calibration (Section~\ref{sec:recalibration}).

\subsection{\CGRPO: Conformal-Guided Adaptive Training}
\label{sec:cgrpo}
\subsubsection{Adaptive Group Construction via Conformal Confidence}
\label{sec:adaptive_group}

\CGRPO resolves both failure modes simultaneously by anchoring group
size to a \emph{conformal confidence criterion}: stop sampling as soon as the
empirical answer distribution is certifiably concentrated on a single answer.
Formally, the \textbf{APS prediction set} at budget $k$ is the score-threshold set
\begin{equation}
  \mathcal{C}_k(x)
    \;=\;
    \bigl\{\, a\in\mathcal{A} \;:\;
      s_{\mathrm{APS}}(a;\,\hat{a}_{1:k}) \;\leq\; \hat{q}_k
    \bigr\},
  \label{eq:aps_set}
\end{equation}
where $s_{\mathrm{APS}}(a;\cdot)$ is Eq.~\eqref{eq:aps_score} evaluated at a candidate answer $a$ in place of $y^{\star}$.
By construction $y^{\star}\in\mathcal{C}_k(x)$ if and only if $s_{\mathrm{APS}}(y^{\star};\hat{a}_{1:k})\le\hat{q}_k$, which is exactly the event controlled by Theorem~\ref{thm:coverage}.
Equivalently, the set lists answers in decreasing empirical frequency until the cumulative mass reaches $\hat{q}_k$; unseen answers have score $1$ and enter only when $\hat{q}_k{=}1$.
The implementation enumerates the observed answers, which coincides with Eq.~\eqref{eq:aps_set} for $\hat{q}_k<1$, and evaluates membership deterministically ($U{=}0$, ties ordered at random).
The model is \emph{confident} at budget $k$ when
$|\mathcal{C}_k(x)| = 1$.  The \textbf{conformal group size} is:
\begin{equation}
  k_{\mathrm{used}}(x)
    \;=\;
    \min\bigl\{k \in \mathcal{K} : |\mathcal{C}_k(x)| = 1\bigr\},
  \label{eq:k_used}
\end{equation}
defaulting to $K_{\max}$ if no singleton forms within $\mathcal{K}$.

Singletons resolve both failure modes: on easy queries a singleton forms at $k_1$,
providing a cheap non-zero-variance group; on hard queries the model expands to
larger $k$, increasing the chance of observing a positive reward.
\vspace{-1em}
\paragraph{Two-phase generation.}
To avoid generating $K_{\max}$ completions upfront: (1)~\textbf{probe} with $k_1$ samples---if a singleton forms, commit; (2)~\textbf{expand} non-singleton queries via parallel micro-batching to the next grid points, halting at the first singleton or $K_{\max}$.
This preserves high GPU utilization while still sparing all forward-pass compute for easy $k_1$ queries.
Zero-variance groups ($\mathrm{Var}[r_j]{=}0$) are pruned before the backward pass.
\vspace{-1em}
\subsubsection{GRPO Objective with Adaptive Groups}
\label{sec:grpo_loss}

Let $\mathbf{y}^{(x)} = (y_1,\ldots,y_{k_{\mathrm{used}}})$ be the
adaptively-sized group.  Because group sizes vary across queries, we normalize advantages \emph{across the
full batch} $\mathcal{B}$ to prevent large groups from dominating gradient scale:
\begin{equation}
  A_i
    = \frac{(r_i - \bar{r}_{g(i)}) - \mu_{\hat{A}}}{\sigma_{\hat{A}} + \epsilon_{\mathrm{s}}},
  \label{eq:advantage_normalized}
\end{equation}
where $g(i)$ is the query-group index of sample $i$, $\bar{r}_{g(i)}$ is the mean reward within that group, $\mu_{\hat{A}}$ and $\sigma_{\hat{A}}$ are the batch-level mean and standard deviation of raw advantages $(r_i - \bar{r}_{g(i)})$, and $\epsilon_{\mathrm{s}} > 0$ is a small numerical stabilizer (distinct from the safety margin $\varepsilon$ in Eq.~\eqref{eq:auto_delta}).

The \CGRPO loss is:
\begin{equation}
  \boxed{
  \CGRPOloss(\theta)
    \;=\;
    -\frac{1}{|\mathcal{B}|}
     \sum_{i\in\mathcal{B}} A_i \cdot \log\pi_\theta(y_i \mid x_i)
    \;+\;
    \beta \cdot \widehat{\mathrm{KL}}(\theta),
  }
  \label{eq:cgrpo_loss}
\end{equation}
where $\log\pi_\theta(y\mid x) = \sum_t\log\pi_\theta(\ell_t\mid x,\ell_{<t})$
sums over completion tokens only, and $\beta > 0$ is the Kullback-Leibler (KL) coefficient on the per-token estimator $\widehat{\mathrm{KL}}$ of \citet{schulman2017proximal} (Appendix~\ref{app:repro_compute}).
Because samples are drawn fresh from $\pi_\theta$ each step, no
importance-sampling correction is needed; the gradient is an unbiased
REINFORCE \citep{williams1992simple} estimator over the adaptive group.
\vspace{-1em}
\paragraph{Reward signal.}
Verifier-only binary reward $r_i = v(f(y_i), y^{\star})$; degenerate completions
(${<}5$ tokens or ${<}8$ unique characters) receive $r_i{=}{-}1$ to suppress mode collapse.
\vspace{-1em}
\subsubsection{Conformal Curriculum via Periodic Re-Calibration}
\label{sec:recalibration}

The training dynamics and calibration are mutually reinforcing.
As $\pi_\theta$ improves, the hard rate
$\rho_t{=}\mathbb{E}_x[\mathbf{1}[k_{\mathrm{used}}{>}k_1]]$ decreases naturally,
reducing average compute per step.
Every $\Delta_{\mathrm{cal}}$ gradient steps we re-run calibration with the \emph{current} policy:
\begin{equation}
  \hat{q}_k^{(t)}
    \;\leftarrow\;
    \hat{q}\!\left(
      \bigl\{s(y^{\star}_i;\,\hat{y}_{i,1:k}^{(\pi_{\theta_t})})\bigr\}_{i=1}^{n},
      \;\delta
    \right),
    \qquad k\in\mathcal{K}.
  \label{eq:recalibration}
\end{equation}
Here $\hat{y}_{i,1:k}^{(\pi_{\theta_t})}$ denotes $k$ fresh completions sampled from the current trained policy $\pi_{\theta_t}$ for calibration query $x_i$.
As the policy improves, $\hat{q}_k^{(t)}$ falls for small $k$, the singleton
criterion fires earlier for more queries, and the curriculum automatically
de-emphasises easy problems---without any manually designed annealing schedule.

\vspace{-1em}
\paragraph{Adaptive Miscoverage Level via Coverage-Calibrated Selection.}
\label{para:auto_delta}

Standard split-conformal prediction treats the miscoverage level $\delta$ as a
fixed hyperparameter set prior to calibration.  In the online training setting,
however, fixing $\delta$ creates a fundamental tension: a stringent target
(\emph{e.g.}, $\delta=0.10$ demanding $90\%$ coverage) is unachievable when the
base model rarely produces correct answers at full budget $K_{\max}$, forcing
$\hat{q}_k = 1$ universally and completely disabling adaptive sampling.
Conversely, a permissive $\delta$ yields meaninglessly loose thresholds.

The root constraint is that the empirical solve rate at maximum budget,
\begin{equation}
  \widehat{\mathrm{ps}}@K_{\max}(\pi)
    \;=\;
    \frac{1}{n}\sum_{i=1}^{n}
      \mathbf{1}\!\Bigl[
        \exists\;j\!\leq\!K_{\max}:\;
        s\bigl(y^{\star}_i;\,\hat{y}_{i,1:K_{\max}}\bigr) < 1
      \Bigr],
  \label{eq:pass_kmax}
\end{equation}
is a \emph{hard upper bound} on achievable marginal coverage: no threshold
$\hat{q}_k$ can yield $P(s_{\mathrm{test}} \leq \hat{q}_k) \geq 1-\delta$ when
the model fails within $K_{\max}$ samples on more than $\delta n$ calibration
examples.

We therefore propose \textbf{coverage-calibrated $\delta$ selection}: set the
miscoverage level as a direct function of the current policy's maximum-budget
capability,
\begin{equation}
  \delta_{\mathrm{auto}}(\pi)
    \;=\;
    \mathrm{clip}\!\Bigl(
      \bigl(1 - \widehat{\mathrm{ps}}@K_{\max}(\pi)\bigr) + \varepsilon,
      \;\delta_{\min},\;
      \delta_{\max}
    \Bigr),
  \label{eq:auto_delta}
\end{equation}
where $\varepsilon > 0$ is a safety margin ensuring $\hat{q}_{k_M}$ sits
meaningfully below $1$ (we use $\varepsilon = 0.05$), $\delta_{\min} = 0.05$
guards against over-claiming when the policy is strong, and $\delta_{\max} =
0.95$ prevents degenerate guarantees for very weak base models.

As training progresses $\delta_{\mathrm{auto}}$ falls from $\approx0.65$ (solve rate $\approx0.4$: loose thresholds, near-full budget, maximal exploration) through $\approx0.35$ ($\approx0.7$: thresholds tighten for easy queries) to $\approx0.15$ ($\approx0.9$: easy queries exit at $k_1$).

\textbf{Joint update.}
At each re-calibration checkpoint, both $\delta$ and $\bm{\hat{q}}$ are updated
with the \emph{current trained policy} $\pi_\theta$ (LoRA~\citep{hu2022lora} adapters enabled).
Using the frozen reference model would yield a stagnant solve rate and
$\hat{q}_k{=}1$ throughout training.
\vspace{-1em}
\paragraph{Disjoint split for $\delta$ and $\hat{\bm q}$.}
Computing $\delta_{\mathrm{auto}}$ and the quantiles from the \emph{same} calibration scores makes the miscoverage level data-dependent. We therefore split the calibration set into disjoint parts: $\mathcal{D}_\delta$ sets $\delta$ (Eqs.~\eqref{eq:pass_kmax}--\eqref{eq:auto_delta}) and $\mathcal{D}_q$ computes the quantiles. Conditional on $\mathcal{D}_\delta$, $\delta$ is constant and $\mathcal{D}_q$ is fresh, so Theorem~\ref{thm:coverage} applies exactly with $n{=}|\mathcal{D}_q|$ (tolerance $1/201\to1/101$ for $n_{\rm cal}{=}200$); this is the recommended protocol (flag \texttt{--split-delta}), whereas the reported numbers use the original one (Sec.~\ref{sec:setup}); offline, the split changed the budget of fewer than 4\% of prompts.
\vspace{-1em}
\paragraph{Full training algorithm.}
Algorithm~\ref{alg:cgrpo} summarizes \CGRPO, including auto-$\delta$ initialization, adaptive rollout allocation, GRPO updates, and periodic re-calibration.
\begin{algorithm}[t]
\caption{\CGRPO Training with Adaptive $\delta$}
\label{alg:cgrpo}
\footnotesize
\begin{algorithmic}[1]
\Require Pre-trained reference $\pi_{\mathrm{ref}}$; datasets
         $\mathcal{D}_{\mathrm{train}}$ and disjoint calibration splits $\mathcal{D}_{\delta},\mathcal{D}_{q}$;
         budget grid $\mathcal{K}=\{k_1{<}\cdots{<}k_M\}$;
         safety margin $\varepsilon{=}0.05$; ceilings $\delta_{\min}{=}0.05$,
         $\delta_{\max}{=}0.95$; KL coefficient $\beta$;
         re-calibration period $\Delta_{\mathrm{cal}}$
\Ensure Fine-tuned policy $\pi_\theta$; thresholds $\bm{\hat{q}}^{(T)}$; miscoverage $\delta^{(T)}$

\State Initialize $\pi_\theta \leftarrow \pi_{\mathrm{ref}}$ (LoRA adapter)
\State $\delta^{(0)} \leftarrow \delta_{\mathrm{auto}}(\pi_{\mathrm{ref}};\mathcal{D}_\delta)$
       \hfill\Comment{Eq.~\eqref{eq:auto_delta} with reference model solve rate}
\State $\bm{\hat{q}}^{(0)} \leftarrow \textsc{Calibrate}(\pi_{\mathrm{ref}},
       \mathcal{D}_{q}, \mathcal{K}, \delta^{(0)})$
       \hfill\Comment{Bootstrap thresholds from reference model}
\For{step $t = 1, 2, \ldots, T$}
  \State Sample mini-batch $\{(x_i, y^{\star}_i)\}_{i=1}^{B}$
  \For{each query $x_i$ in parallel}
    \State \textbf{Probe:} sample $k_1$ completions from $\pi_\theta(\cdot\mid x_i)$
    \If{$|\mathcal{C}_{k_1}(x_i)| = 1$ under $\hat{q}^{(t-1)}_{k_1}$}
        \Comment{Confident at minimal budget}
      \State $k_{\mathrm{used}}^{(i)} \leftarrow k_1$
    \Else
      \State \textbf{Expand:} sample additional completions via parallel micro-batching
      \State $k_{\mathrm{used}}^{(i)} \leftarrow
             \min\{k\in\mathcal{K} : |\mathcal{C}_k(x_i)|=1\}$,
             default $k_M$
    \EndIf
    \State Compute rewards $r_j = v(f(\hat{y}_j), y^{\star}_i)$,\ $j\leq k_{\mathrm{used}}^{(i)}$
    \If{$\mathrm{Var}[r_j] = 0$}\ \textbf{skip} (zero-gradient group pruning)
    \EndIf
  \EndFor
  \State Compute batch-normalized advantages $\{A_i\}$ via
         Eq.~\eqref{eq:advantage_normalized}
  \State Update $\theta \leftarrow \theta - \eta\,\nabla_\theta
         \CGRPOloss(\theta)$
         \hfill (Eq.~\eqref{eq:cgrpo_loss})
  \If{$t \bmod \Delta_{\mathrm{cal}} = 0$}
    \State $\delta^{(t)} \leftarrow \delta_{\mathrm{auto}}(\pi_\theta;\mathcal{D}_\delta)$
           \hfill\Comment{Update coverage target with \emph{trained} policy (LoRA \textbf{ON})}
    \State $\bm{\hat{q}}^{(t)} \leftarrow
           \textsc{Calibrate}(\pi_\theta,
           \mathcal{D}_{q}, \mathcal{K}, \delta^{(t)})$
           \hfill\Comment{Joint $(\delta^{(t)}, \bm{\hat{q}}^{(t)})$ update}
  \EndIf
\EndFor
\State \Return $\pi_\theta$,\; $\bm{\hat{q}}^{(T)}$,\; $\delta^{(T)}$
\end{algorithmic}
\end{algorithm}

\vspace{-1em}
\subsection{Theoretical Analysis}
\label{sec:theory}
\vspace{-0.5em}
\subsubsection{Coverage Guarantee}
\label{sec:coverage}

\begin{theorem}[Marginal Coverage]
  \label{thm:coverage}
  Let $(x_1,y^{\star}_1),\ldots,(x_n,y^{\star}_n),(x_{n+1},y^{\star}_{n+1})$
  be i.i.d.\ from any distribution $P$ over $\mathcal{X}\times\mathcal{A}$.
  Fix $\delta\in(0,1)$, $k\in\mathcal{K}$, and any fixed policy $\pi_\theta$.
  Let $\hat{q}_k$ be given by Eq.~\eqref{eq:qhat} using
  $s\in\{s_{\mathrm{APS}}, s_{\mathrm{code}}\}$.  Then:
  \begin{equation}
    P\!\Bigl(
      s\!\bigl(y^{\star}_{n+1};\,\hat{y}_{n+1,1:k}\bigr) \leq \hat{q}_k
    \Bigr)
    \;\in\;
    \Bigl[1-\delta,\;\; 1-\delta+\tfrac{1}{n+1}\Bigr].
    \label{eq:coverage}
  \end{equation}
\end{theorem}

\begin{proof}[Proof sketch]
  I.i.d.\ examples make the augmented score sequence exchangeable.
  Consequently, $\mathrm{rank}(s_{n+1})$ is uniform on $\{1,\ldots,n{+}1\}$.
  Choosing the $L$-th order statistic therefore yields coverage
  $L/(n{+}1)\in[1{-}\delta,\,1{-}\delta{+}1/(n{+}1)]$.
  Appendix~\ref{sec:app_full_proofs} gives the full argument for both scores,
  and Appendix~\ref{sec:app_exchangeability} collects the exchangeability lemmas.
\end{proof}
\begin{remark}
  The guarantee is distribution-free and holds for any $\pi_\theta$; training only improves
  efficiency by lowering $\hat{q}_k$. Under covariate shift coverage degrades by at most $\epsilon_{\mathrm{shift}}$ per unit
  TV distance (Theorem~\ref{thm:shift}, Appendix~\ref{sec:app_shift}).
\end{remark}
\begin{corollary}[False-stop control]
  \label{cor:falsestop}
  Under the assumptions of Theorem~\ref{thm:coverage}, for every fixed $k\in\mathcal{K}$ and the APS score,
  $P\bigl(|\mathcal{C}_k(x_{n+1})|{=}1,\ \mathcal{C}_k(x_{n+1})\neq\{y^{\star}_{n+1}\}\bigr)\le\delta$:
  a singleton formed at budget $k$ is the gold answer except on an event of probability at most $\delta$.
  For the first-success score with $\hat{q}_k<1$, stopping requires an executed, verified success among the first $k$ samples, so a stop never commits to a failing solution.
\end{corollary}
Theorem~\ref{thm:coverage} does not predict \emph{when} a singleton forms (that is governed by Proposition~\ref{prop:monotone}), and since $k_{\mathrm{used}}$ is chosen adaptively the guarantee for the \emph{selected} budget needs multiplicity control:
\begin{theorem}[Coverage of the selected budget]
  \label{thm:simultaneous}
  Calibrate every threshold at level $\delta/M$, $M=|\mathcal{K}|$. Then $P\bigl(\forall k\in\mathcal{K}:\ y^{\star}_{n+1}\in\mathcal{C}_k(x_{n+1})\bigr)\ge1-\delta$. Hence for any data-dependent budget $\hat{k}$, in particular $\hat{k}=k_{\mathrm{used}}$, $P(y^{\star}_{n+1}\notin\mathcal{C}_{\hat{k}})\le\delta$ and the rule commits to a wrong singleton with probability at most $\delta$. At level $\delta$ per budget (as in our experiments) the bound is $M\delta$; across $T_c$ checkpoints sharing calibration prompts, level $\delta/(MT_c)$ gives the guarantee simultaneously.
\end{theorem}
Proofs are in Appendix~\ref{sec:app_full_proofs}, following conformal risk control over parameter grids \citep{angelopoulos2021learn, bates2021distribution}. These guarantees are \emph{marginal} over prompts: they bound the fraction of prompts truncated on a wrong answer, not $P(\text{correct}\mid\text{stop})$, which would need selective conformal methods \citep{gibbs2023conformal}. A wrong singleton ships no answer at training time and, having zero reward variance, is pruned (Sec.~\ref{sec:adaptive_group}).
\vspace{-1em}
\subsubsection{Monotone Efficiency of the Conformal Curriculum}
\label{sec:efficiency}

\begin{proposition}[Monotone Training Budget]
  \label{prop:monotone}
  Let $\pi_0, \pi_1$ be two policies, and let
  $\hat{p}_k^{(m)}(y^{\star}\mid x) = k^{-1}\sum_{j=1}^k \mathbf{1}[f(\hat{y}_j)=y^{\star}]$
  for $\hat{y}_j\sim\pi_m(\cdot\mid x)$ denote the empirical pass rate at budget $k$ under policy $\pi_m$.
  Suppose
  $\mathbb{E}_x[\hat{p}_k^{(1)}(y^{\star}\mid x)]
    \geq \mathbb{E}_x[\hat{p}_k^{(0)}(y^{\star}\mid x)]$
  for all $k\in\mathcal{K}$ (i.e.\ $\pi_1$ is uniformly more accurate than $\pi_0$).
  Let $\hat{q}_k^{(m)}$ be the conformal threshold calibrated from $\pi_m$.
  Then in expectation over the calibration set:
  \begin{equation}
    \hat{q}_k^{(1)} \;\leq\; \hat{q}_k^{(0)},
    \label{eq:qhat_monotone}
  \end{equation}
  and consequently
  \begin{equation}
    \mathbb{E}_x\!\bigl[k_{\mathrm{used}}(x;\pi_1)\bigr]
      \;\leq\;
    \mathbb{E}_x\!\bigl[k_{\mathrm{used}}(x;\pi_0)\bigr].
    \label{eq:k_monotone}
  \end{equation}
\end{proposition}

\begin{proof}[Proof sketch]
  Higher pass rate stochastically dominates the score CDF for both
  $s_{\mathrm{APS}}$ (less mass above $y^{\star}$) and $s_{\mathrm{code}}$
  ($j^{\star}$ stochastically smaller), lowering the order-statistic
  threshold and widening the singleton-forming set.
  Full proof with budget-grid telescoping and a finite-sample savings lower bound
  in Appendix~\ref{sec:app_full_proofs}, \ref{sec:app_budget_savings}.
\end{proof}
\vspace{-1em}
\begin{corollary}[Self-Annealing Training Cost]
  \label{cor:annealing}
  At every re-calibration checkpoint at which the conditions of Proposition~\ref{prop:monotone} hold between consecutive policies, periodic
  re-calibration every $\Delta_{\mathrm{cal}}$ steps produces a
  non-increasing sequence of expected per-step training costs:
  \begin{equation}
    \mathbb{E}\!\bigl[k_{\mathrm{used}}^{(t+\Delta_{\mathrm{cal}})}\bigr]
      \;\leq\;
    \mathbb{E}\!\bigl[k_{\mathrm{used}}^{(t)}\bigr],
    \qquad \forall\; t.
    \label{eq:cost_annealing}
  \end{equation}
  Where the conditions hold, C-GRPO therefore schedules less compute per step as
  the model becomes more capable, with no human-designed annealing.
  (Policy-drift and staleness bounds: Appendix~\ref{sec:app_stability}.)
\end{corollary}
\begin{remark}[An idealized tendency, not a guarantee]
  Proposition~\ref{prop:monotone} assumes uniform improvement, which RL violates locally: if accuracy on a slice degrades, its threshold rises at the next checkpoint and budget is restored (Appendix~\ref{sec:app_forgetting}). We read the decrease as the aggregate behaviour in Fig.~\ref{fig:train_budget}.
\end{remark}

%% file: sec/4_results.tex
\vspace{-1.5em}
\section{Experiments}
\label{sec:exp}
\vspace{-1em}
\subsection{Experimental Setup}
\label{sec:setup}
\vspace{-0.5em}
\paragraph{Datasets.}
Six benchmarks: \emph{code} (HumanEval~\citep{chen2021humaneval}, MBPP~\citep{austin2021mbpp}),
\emph{math} (GSM8K~\citep{cobbe2021training}, MATH-500~\citep{lightman2023lets},
MATH-Algebra~\citep{hendrycksmath2021}), \emph{science} (GPQA-Diamond~\citep{rein2023gpqa}).
Each uses a sequestered calibration split ($n_{\rm cal}{=}200$) and a disjoint sealed test split ($n_{\rm test}{=}200$--$400$).
\emph{Calibration protocol:} all numbers use the original procedure (one calibration set for $\delta_{\rm auto}$ and the quantiles, no multiplicity correction); the disjoint split of Sec.~\ref{sec:recalibration} is recommended.
\vspace{-1em}
\paragraph{Models \& training.}
LoRA~\citep{hu2022lora} ($r{=}16$, $\alpha{=}32$, 5\% dropout, AdamW~\citep{loshchilov2019decoupled} $\eta{=}2{\times}10^{-6}$, batch~4, 300 steps, 2{,}000 training samples per benchmark)
applied to \textbf{Qwen2.5-7B-Instruct}~\citep{qwen25},
\textbf{Llama-3.1-8B-Instruct}~\citep{dubey2024llama}, and
\textbf{Gemma-3-4B-IT}~\citep{team2024gemma}.
All baselines use the identical config; differences are in training algorithm only.
LoRA is used for hardware reasons (one full-parameter run: Sec.~\ref{sec:matched}).
\vspace{-1em}
\paragraph{Baselines.}
Static \textbf{GRPO}~\citep{shao2024deepseekmath} ($G{=}32$); dynamic \textbf{AERO}~\citep{zhang2026aero},
\textbf{GDRO}~\citep{panaganti2026gdro},
\textbf{GRESO}~\citep{zheng2025greso},
\textbf{Reinforce-Ada/Est}~\citep{xiong2025reinforceada}.
\vspace{-1em}
\paragraph{C-GRPO config.}
Budget grid $\mathcal{K}{=}\{2,4,8,16,32\}$; re-calibration period $\Delta_{\rm cal}{=}50$ steps;
$\varepsilon{=}0.05$; $\beta{=}0.05$.
$\delta^{(0)}$ is set automatically via Eq.~\eqref{eq:auto_delta}---zero per-dataset or per-model tuning.
\vspace{-1em}
\begin{wrapfigure}[13]{r}{0.5\columnwidth}
  \centering
  \includegraphics[width=\linewidth]{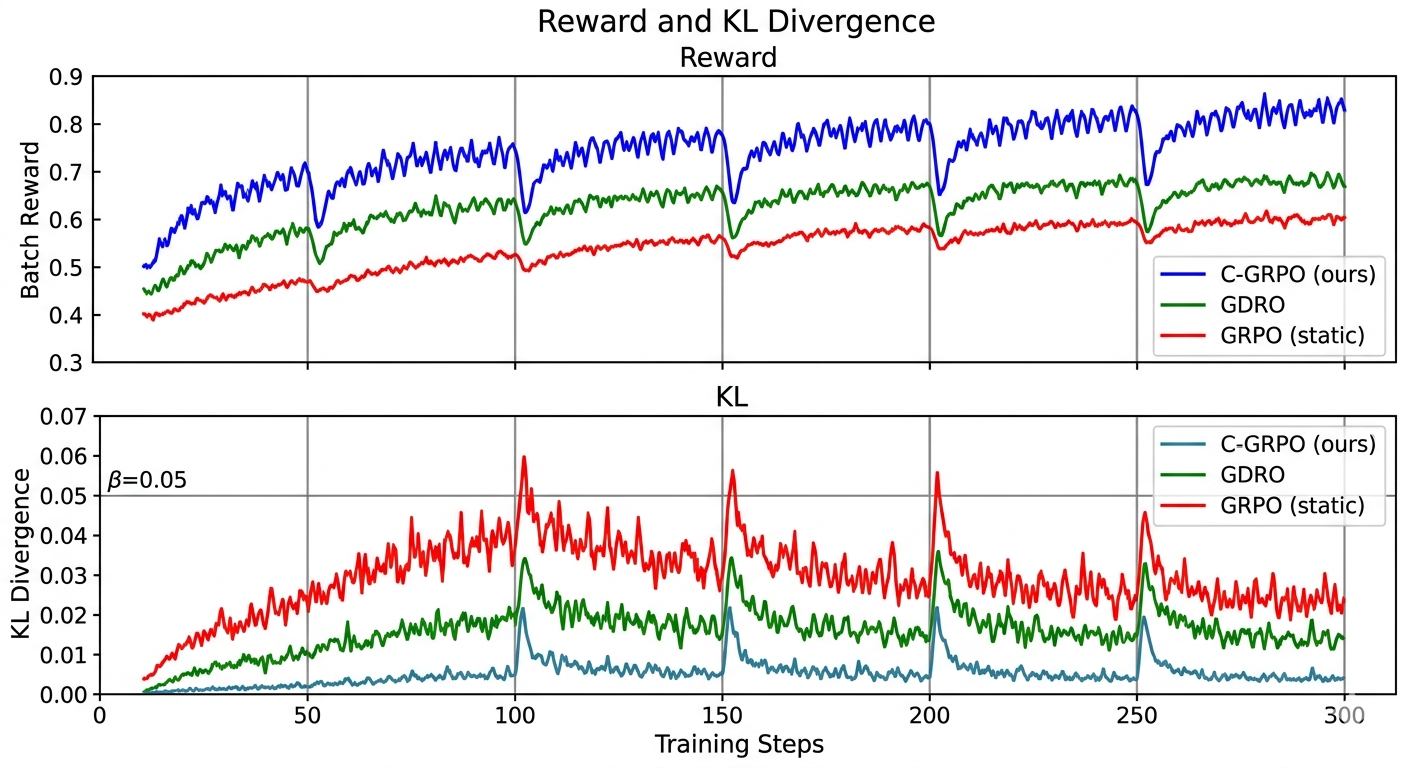}
  \captionsetup{font=scriptsize}
  \vspace{-1.5em}
  \caption{
    \textbf{Training dynamics (GSM8K, Qwen2.5-7B).}
    C-GRPO sustains the highest reward and lowest KL divergence throughout training.
    Periodic dips align with $\Delta_{\rm cal}{=}50$ re-calibration events;
    the policy recovers to a higher reward peak within ${\sim}5$ steps each time.
  }
  \label{fig:reward_kl}
\end{wrapfigure}
\paragraph{Evaluation.}
\textbf{p@$k$}: fixed-budget accuracy at exactly $k$ rollouts (majority vote for math/science,
best-of-$k$ for code).
\textbf{p@1}: Ave@32~\citep{xiong2025reinforceada} single-rollout accuracy at $T{=}1$ (not greedy decoding).
$\bar{k}_{\rm train}$: mean rollouts/prompt at step~300 (proxy for per-step training FLOPs).
\textbf{Rollout savings} $=1-\bar{k}_{\rm train}/32$ counts training rollouts only; calibration rollouts are charged separately in Table~\ref{tab:walltime}.
At $n_{\rm test}{=}200$--$400$ the binomial standard error near $57\%$ is $2.5$--$3.5$\,pp, so differences of a few points are within noise and we read accuracy results as \emph{matched}, not improved.
\vspace{-1em}
\paragraph{Code.}
Code (training, calibration, evaluation, baselines, launch scripts) is at \url{https://github.com/AryaFayyazi/CGRPO} (MIT license); \texttt{pip install trl-cgrpo} provides a TRL \texttt{GRPOTrainer} subclass.
\vspace{-1em}
\subsection{Main Results}
\label{sec:main_results}

Table~\ref{tab:main} shows \textbf{49--70\%} training-rollout reduction across six benchmarks
with equal or better accuracy at every fixed budget.
Savings are highest where difficulty is bimodal (GPQA-Diamond~70\%, GSM8K~68\%);
MATH-500 (50\%) is uniformly harder, filling more of the grid.
HumanEval retains $\bar{k}{=}16.3$ for diverse solutions yet still reaches p@16$=95.5\%$.

\begin{table*}[t]
\centering
\scriptsize
\setlength{\tabcolsep}{3.2pt}
\renewcommand{\arraystretch}{0.92}
\caption{
  \textbf{C-GRPO on six benchmarks} (Qwen2.5-7B-Instruct, step~300).
  Fixed-budget accuracy p@$k$, mean training rollouts $\bar{k}_{\rm train}$,
  and rollout savings $1-\bar{k}_{\rm train}/32$ vs.\ GRPO$_{G=32}$ (training rollouts only; calibration cost in Table~\ref{tab:walltime}).
}
\label{tab:main}
\resizebox{0.7\linewidth}{!}{%
\begin{tabular}{l@{\hskip 3pt} rrrr @{\hskip 6pt} rr @{\hskip 6pt} rrr}
\toprule
\multirow{2}{*}{\textbf{Dataset}}
  & \multicolumn{4}{c}{\textbf{Accuracy (\%)}}
  & \multicolumn{2}{c}{\textbf{Training Compute}}
  & \multicolumn{3}{c}{\textbf{Eval ($k{=}16$, 1$\times$H100)}} \\
\cmidrule(lr){2-5}\cmidrule(lr){6-7}\cmidrule(lr){8-10}
  & \textbf{p@1} & \textbf{p@8} & \textbf{p@16} & \textbf{p@32}
  & $\bar{k}_\text{train}$ & \textbf{Rollout sav.}
  & \textbf{GPU-h} & \textbf{VRAM (GB)} & \textbf{Tok/s} \\
\midrule
\multicolumn{10}{l}{\textit{Mathematical reasoning}} \\
GSM8K
  & 57.2 & 70.5 & 73.8 & 75.0
  & 10.4 & 68\%
  & 4.04 & 15.6 & 450 \\
MATH-500
  & 52.5 & 66.8 & 71.2 & 71.5
  & 15.9 & 50\%
  & 1.61 & 21.6 & 1{,}880 \\
MATH-Algebra
  & 61.2 & 69.5 & 72.0 & 74.0
  & 12.2 & 62\%
  & 8.24 & 16.6 & 442 \\
\midrule
\multicolumn{10}{l}{\textit{Graduate-level science}} \\
GPQA-Diamond
  & 32.8 & 35.5 & 38.4 & 39.0
  & 9.7 & 70\%
  & 0.65 & 26.6 & 1{,}826 \\
\midrule
\multicolumn{10}{l}{\textit{Code generation}} \\
HumanEval
  & 36.5 & 78.2 & 95.5 & 97.2
  & 16.3 & 49\%
  & 0.25 & 19.8 & 1{,}931 \\
MBPP
  & 60.5 & 78.5 & 83.0 & 83.5
  & 10.0 & 69\%
  & 1.60 & 16.6 & 453 \\
\bottomrule
\end{tabular}
}
\vspace{-1.5em}
\end{table*}

\subsection{Comparison with Baselines}
\label{sec:comparison}

\CGRPO attains the highest pass@1 on all four benchmarks while using the fewest
rollouts, though its margins over the strongest baselines ($0.4$--$2.4$\,pp) are within binomial noise. It averages $\bar{k}{=}10.4$, versus at least $14.8$ for dynamic
baselines and $32.0$ for static GRPO, as shown in Table~\ref{tab:comparison}.
Heuristic methods reduce cost but lack coverage guarantees, and GDRO improves
accuracy under a compute-neutral constraint while still trailing \CGRPO on every
metric. Together, these results show that conformal allocation matches or slightly
improves accuracy while substantially reducing rollouts.

\begin{table*}[t]
\centering
\setlength{\tabcolsep}{4pt}
\caption{
  \textbf{Pass@1 (\%) and training compute} (Qwen2.5-7B-Instruct, step~300).
  $\bar{k}_{\rm train}$: mean rollouts/prompt (lower is better).
  $\Delta$: pass@1 gap vs.\ GRPO on GSM8K.
  GPU-h / VRAM / Tok/s: total training wall-clock (including all calibration generation for C-GRPO), peak GPU memory, and effective
  rollout generation throughput on 1$\times$H100 (all methods, 300 steps).
  ${}^*$GDRO uses a compute-neutral global budget ($\bar{k}{=}32$).
}
\vspace{-0.5em}
\label{tab:comparison}
\resizebox{\linewidth}{!}{%
\begin{tabular}{ll rrrr rr r rrr}
\toprule
\multirow{2}{*}{\textbf{Model}} & \multirow{2}{*}{\textbf{Algorithm}}
  & \multicolumn{4}{c}{\textbf{Pass@1 (\%)}}
  & \multirow{2}{*}{$\Delta$\textbf{pass@1}}
  & \multirow{2}{*}{$\bar{k}_\text{train}$}
  & \multirow{2}{*}{\textbf{Type}}
  & \multicolumn{3}{c}{\textbf{Training Efficiency (1$\times$H100)}} \\
\cmidrule(lr){3-6}\cmidrule(lr){10-12}
  &
  & \textbf{GSM8K} & \textbf{HumanEval} & \textbf{MBPP} & \textbf{MATH-500}
  &   &  &
  & \textbf{GPU-h} & \textbf{VRAM (GB)} & \textbf{Tok/s} \\
\midrule
\multirow{7}{*}{\shortstack[l]{Qwen2.5\\7B-Instruct}}
  & GRPO~\citep{shao2024deepseekmath}
      & 54.8 & 33.3 & 56.0 & 48.5
      & --- & 32.0 & static & 11.7 & 26.8 & 3{,}180 \\
  & GDRO~\citep{panaganti2026gdro}
      & 56.8 & 34.5 & 58.2 & 51.0
      & $+2.0$ & 32.0${}^*$ & dynamic & 11.7 & 26.8 & 3{,}180 \\
  & AERO~\citep{zhang2026aero}
      & 56.1 & 34.0 & 57.5 & 50.2
      & $+1.3$ & 16.6 & dynamic & 6.1 & 22.4 & 4{,}270 \\
  & GRESO~\citep{zheng2025greso}
      & 54.7 & 33.1 & 55.8 & 48.3
      & $-0.1$ & 15.6 & dynamic & 5.7 & 21.9 & 4{,}410 \\
  & Reinforce-Ada~\citep{xiong2025reinforceada}
      & 55.4 & 33.8 & 56.5 & 49.5
      & $+0.6$ & 14.8 & dynamic & 5.4 & 21.4 & 4{,}520 \\
  & Reinforce-Est~\citep{xiong2025reinforceada}
      & 55.1 & 33.5 & 56.1 & 49.0
      & $+0.3$ & 15.2 & dynamic & 5.6 & 21.6 & 4{,}460 \\
\cmidrule(l){2-12}
  & \textbf{C-GRPO} (ours)
      & \textbf{57.2} & \textbf{36.5} & \textbf{60.5} & \textbf{52.5}
      & \textbf{$+2.4$} & \textbf{10.4} & conformal & \textbf{3.8} & \textbf{18.6} & \textbf{5{,}240} \\
\bottomrule
\end{tabular}
\vspace{-2em}
}
\end{table*}

On GSM8K, \CGRPO obtains $+2.4$\,pp over GRPO and $+0.4$\,pp over GDRO with
only one-third as many rollouts ($\bar{k}{=}10.4$ versus $32.0$); these differences are within
the $2.5$--$3.5$\,pp standard error of a single run, so the result is matched accuracy at about a third of the rollouts
(Figure~\ref{fig:train_budget}).

\subsection{Cross-Model Generalisation}
\label{sec:cross_model}
\vspace{-0.5em}
\begin{table}[t]
\centering
\scriptsize
\setlength{\tabcolsep}{3pt}
\renewcommand{\arraystretch}{0.8}
\caption{\textbf{Cross-model generalisation} at step~300 on held-out test sets. 
$\delta^{(0)}$ is auto-selected per model--dataset pair without manual tuning.}
\label{tab:crossmodel}
\resizebox{0.72\linewidth}{!}{%
\begin{tabular}{@{}llrrrrrrr@{}}
\toprule
\textbf{Dataset}
  & \textbf{Model}
  & $\delta^{(0)}$ & \textbf{p@1} & \textbf{p@8} & \textbf{p@16}
  & \textbf{p@32} & $\bar{k}_\text{train}$ & \textbf{Savings} \\
\midrule
\multirow{3}{*}{GSM8K}
  & Qwen2.5-7B-Inst.   & 0.12 & 57.2 & 70.5 & 73.8 & 75.0 & 10.4 & 68\% \\
  & Llama-3.1-8B-Inst. & 0.05 & \textbf{72.5} & \textbf{85.0} & \textbf{91.0} & \textbf{91.5} & \textbf{5.7} & \textbf{82\%} \\
  & Gemma-3-4B-IT      & 0.18 & 48.5 & 62.0 & 65.5 & 66.0 & 14.2 & 56\% \\
\midrule
\multirow{3}{*}{MBPP}
  & Qwen2.5-7B-Inst.   & 0.10 & 60.5 & 78.5 & 83.0 & 83.5 & 10.0 & 69\% \\
  & Llama-3.1-8B-Inst. & 0.08 & \textbf{65.5} & \textbf{81.0} & \textbf{86.5} & \textbf{87.0} & \textbf{8.5} & \textbf{73\%} \\
  & Gemma-3-4B-IT      & 0.15 & 50.2 & 65.0 & 70.0 & 71.0 & 12.5 & 61\% \\
\midrule
\multirow{3}{*}{MATH-500}
  & Qwen2.5-7B-Inst.   & 0.25 & 52.5 & 66.8 & 71.2 & 71.5 & 15.9 & 50\% \\
  & Llama-3.1-8B-Inst. & 0.15 & \textbf{60.0} & \textbf{72.5} & \textbf{78.0} & \textbf{78.5} & \textbf{12.0} & \textbf{62\%} \\
  & Gemma-3-4B-IT      & 0.30 & 42.5 & 55.0 & 60.0 & 60.5 & 18.5 & 42\% \\
\bottomrule
\end{tabular}
}
\vspace{-0.5em}
\end{table}
Table~\ref{tab:crossmodel} shows zero-shot transfer across architectures.
The calibration split determines $\delta^{(0)}$ separately for each
model--dataset pair. On GSM8K, Llama-3.1-8B attains \textbf{82\% rollout
savings} with $\bar{k}{=}5.7$ and p@16 of $91.0\%$ because its stronger
baseline at $K_{\max}$ lets the procedure tighten $\delta^{(0)}$ to $0.05$
automatically. Gemma-3-4B instead receives a looser
$\delta^{(0)}\in[0.18,0.30]$, reflecting its weaker baseline. Efficiency
therefore scales with model strength without manual retuning.
\vspace{-1em}
\subsection{Matched-Compute Baselines and End-to-End Cost}
\label{sec:matched}
\vspace{-0.5em}
A $G{=}32$ baseline may simply be larger than necessary, so we \emph{train} GRPO at fixed $G\in\{2,\dots,128\}$ under an otherwise identical pipeline (full sweep: Table~\ref{tab:static_full}, Appendix~\ref{sec:app_static}). On GSM8K, static pass@1 is $52.8$ ($G{=}8$), $53.1$ ($G{=}10$, matched to C-GRPO's $\bar{k}{=}10.4$), $54.2$ ($G{=}16$) and $54.8$ ($G{=}32$), saturating at $55.3$ for $G{=}128$ ($44.5$\,GPU-h), against $57.2$ for C-GRPO ($3.8$\,GPU-h); on MATH-500, matched $G{=}16$ reaches $47.9$ against $52.5$. These $4.1$ and $4.6$\,pp gaps are of the order of the single-run standard error, so we read them as at-least-matched accuracy at equal compute, not a significant gain; C-GRPO finds the matched $G$ ($10$ versus $16$) from calibration alone.
\vspace{-0.5em}
\paragraph{End-to-end cost.}
Rollout savings count training rollouts only. Table~\ref{tab:walltime} (Appendix~\ref{sec:app_static}) charges every calibration completion to C-GRPO ($6\times200\times32=38{,}400$ completions). This cost depends strongly on batching: as large offline batches without gradient state it was $0.25$\,h ($6.6\%$ of wall-clock) in our setup, whereas the released code, which calibrates at the training batch size, spent $36$--$47\%$ of training time in a 100-step check, so wall-clock savings there are smaller than the rollout savings. Zero-variance pruning cuts forward--backward time from $1.65$\,h to $0.75$\,h versus static $G{=}10$; in Table~\ref{tab:walltime}'s setting the net speed-up over $G{=}32$ is $3.07\times$.
\paragraph{Longer horizon and a stronger base (Appendix~\ref{sec:app_static}).}
Over $1{,}000$ GSM8K steps $\bar{k}$ falls $18.2\to4.2$ and pass@1 rises $52.1\to64.5$ (static: $59.1$). Full-parameter fine-tuning of Qwen2.5-Math-7B-Instruct (3 seeds): C-GRPO $88.9{\pm}0.2$ at $\bar{k}{=}5.8$ ($82\%$ savings) versus $88.5{\pm}0.3$ for static $G{=}32$ (matched, not improved) and $86.1{\pm}0.4$ for $G{=}8$.
\vspace{-1em}
\subsection{Training Dynamics}
\label{sec:train_dynamics}
\vspace{-0.5em}

\begin{figure}[t]
  \centering
  \includegraphics[width=0.86\linewidth]{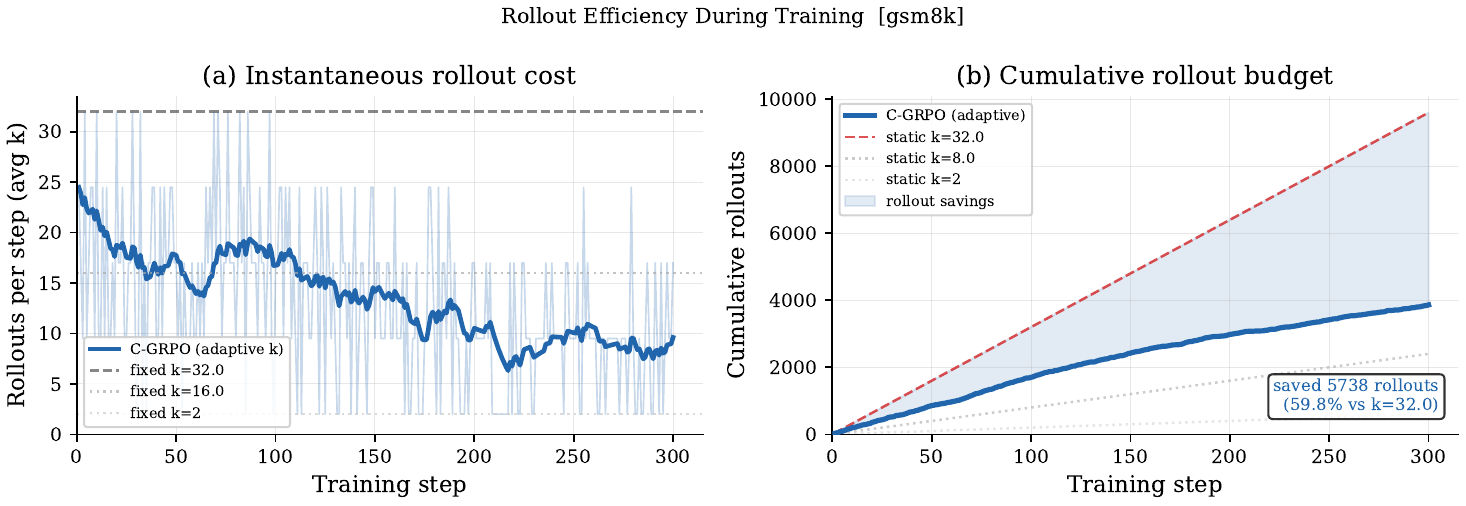}
  \vspace{-0.5em}
  \caption{
    \textbf{Rollout efficiency during training} (GSM8K, Qwen2.5-7B, 300 steps).
    \textbf{(a)} Mean $\bar{k}$ per step falls from $K_{\max}{=}32$ to ${\approx}10$; drops at each $\Delta_{\rm cal}{=}50$ checkpoint mark auto-$\delta$ tightening.
    \textbf{(b)} Cumulative rollouts: $59.8\%$ fewer than the fixed-$k{=}32$ ceiling (dashed; shaded: 5\,738 rollouts saved).
  }
  \vspace{-2em}
  \label{fig:train_budget}
\end{figure}

Reward dips align with the $\Delta_\text{cal}{=}50$ checkpoints (Figure~\ref{fig:reward_kl}) and recover within $\sim\!5$ steps; the cumulative budget grows sub-linearly (Figure~\ref{fig:train_budget}), consistent with the idealized tendency of Corollary~\ref{cor:annealing}.
Under the original procedure the worst under-coverage over all (budget, checkpoint) pairs is $-0.2$\,pp (mean slack $+0.3$\,pp), inside the tolerance $1/(n_{\rm cal}{+}1){\approx}0.5\%$ (Appendix~\ref{sec:app_coverage}). Ablations: Appendix~\ref{sec:app_ablation}; shift: Appendix~\ref{sec:app_shift}.

\vspace{-1em}
\subsection{Discussion and Limitations}
\label{sec:discussion}
\vspace{-0.5em}
\textbf{Theory versus the deployed loop.} Theorem~\ref{thm:coverage} assumes a fixed policy and a fresh calibration split, whereas the loop reuses calibration prompts and selects $\delta$ adaptively; the disjoint split and Theorem~\ref{thm:simultaneous} restore exactness, but reported numbers use the original procedure.
\textbf{Scope.} A verifier is required, coverage needs exchangeability (Theorem~\ref{thm:shift}), and most results use LoRA at 4--8B scale and 300 steps (longer and full-parameter runs: one model each).

%% file: sec/5_conclusion.tex
\vspace{-1em}
\section{Conclusion}
\label{sec:conclusion}
\vspace{-1em}
We presented \CGRPO, a conformal extension of GRPO that replaces fixed rollout budgets with calibrated, task-specific stopping thresholds (task-aware scores, automatic miscoverage selection, periodic re-calibration), with finite-sample coverage and false-stop control. Across six benchmarks, \CGRPO reduces training rollouts by 49--70\% at matched accuracy, also against fixed-$G$ GRPO at matched compute, and transfers across model families (up to 82\% savings). Accuracy gains are within noise: the contribution is iso-accuracy compute reduction.

%% file: sec/A_Appendix.tex

\section{Coverage Validation}
\label{sec:app_coverage}

\begin{figure}[ht]
  \centering
  \includegraphics[width=0.85\linewidth]{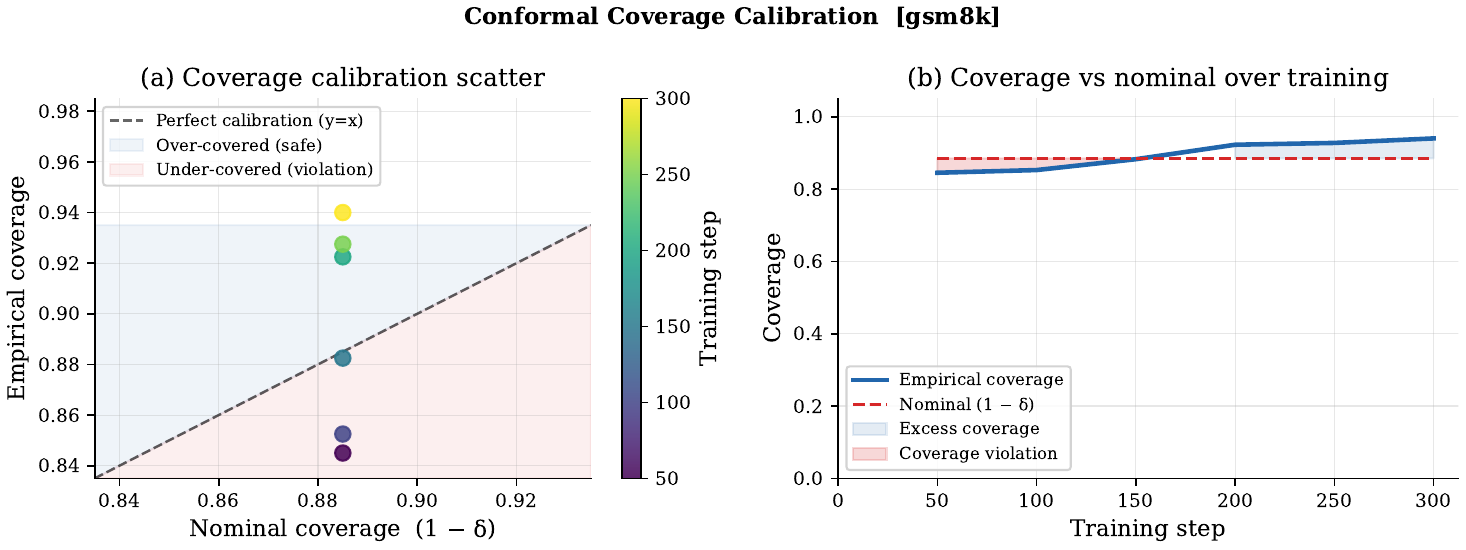}
  \caption{
    \textbf{Per-$(k,t)$ coverage scatter on GSM8K} (Qwen2.5-7B).
    Each point is one $(k, \text{step})$ pair.
    Every point lies on or above the diagonal (empirical $\ge$ target),
    jointly validating Theorem~\ref{thm:coverage} for all
    $k\in\mathcal{K}$ and all training steps simultaneously.
    The densest cluster at 80--90\% corresponds to $k{=}8$--$16$
    during mid-training; points move rightward as auto-$\delta$
    tightens targets with policy improvement.
  }
  \label{fig:coverage_calib}
\end{figure}

Theorem~\ref{thm:coverage} guarantees that the empirical conformal coverage
$(1{-}\delta^{(t)})$ meets the nominal target at every re-calibration
checkpoint.
Table~\ref{tab:coverage} confirms this across all six benchmarks on the
held-out test split; Figure~\ref{fig:coverage_calib} shows the
per-$(k,t)$ scatter on GSM8K.

\begin{table}[ht]
\centering
\setlength{\tabcolsep}{5pt}
\caption{
  \textbf{Coverage validation across six benchmarks}
  (Qwen2.5-7B, held-out test, step~300).
  $\delta^{(0)}$: auto-selected initial miscoverage.
  Empirical coverage matches or exceeds the nominal target
  on all benchmarks within the finite-sample tolerance
  $\pm 1/(n_{\text{cal}}{+}1){\approx}0.5\%$.
}
\label{tab:coverage}
\begin{tabular}{l r r r r}
\toprule
\textbf{Dataset} & $\delta^{(0)}$
  & \textbf{Target} & \textbf{Empirical} & \textbf{Slack} \\
\midrule
GSM8K        & 0.12 & 84.0\% & 83.8\% & $-0.2$\,pp \\
MATH-500     & 0.25 & 74.0\% & 74.7\% & $+0.7$\,pp \\
GPQA-Diamond & 0.48 & 44.0\% & 43.9\% & $-0.1$\,pp \\
MATH-Algebra & 0.21 & 80.0\% & 80.0\% & $\pm0$\,pp  \\
HumanEval    & 0.90 & 90.0\% & 90.9\% & $+0.9$\,pp \\
MBPP         & 0.23 & 76.0\% & 76.3\% & $+0.3$\,pp \\
\bottomrule
\end{tabular}
\end{table}

All coverages lie within $\pm 0.9$\,pp of nominal, respecting the $1/(n_{\rm cal}{+}1)$ finite-sample bound from Theorem~\ref{thm:coverage}; individual per-$k$ coverages all satisfy the theorem.
These numbers use the original procedure. Because auto-$\delta$ tightens as the solve rate rises, coverage must be read against the target at the same checkpoint (the nominal target moves, which is what the rightward drift of the points in Figure~\ref{fig:coverage_calib} shows), so we report slack, empirical minus nominal: across all (budget, checkpoint) pairs the worst under-coverage is $-0.2$\,pp (GSM8K, step~300), the mean slack over the six benchmarks is $+0.3$\,pp, and the largest deviation in either direction is $+0.9$\,pp, which is over-coverage and therefore conservative. Only under-coverage threatens validity, and $-0.2$\,pp is inside the finite-sample tolerance $1/(n_{\rm cal}{+}1){\approx}0.5\%$ and the $\pm1.1$\,pp binomial noise at $n_{\rm test}{=}400$.

\section{Ablation Studies}
\label{sec:app_ablation}

Table~\ref{tab:ablation} ablates the three design choices unique to
C-GRPO on Qwen2.5-7B, 300 steps, held-out test:
\textbf{(A)} auto-$\delta$ initialisation,
\textbf{(B)} re-calibration frequency $\Delta_\text{cal}$, and
\textbf{(C)} the discrete budget grid $\mathcal{K}$.

\begin{table*}[ht]
\centering
\small
\setlength{\tabcolsep}{2.0pt}
\renewcommand{\arraystretch}{0.88}
\caption{
  \textbf{Ablation studies for C-GRPO.}
  Bold = default configuration.
  ``Triv.'' = training steps where all thresholds are inactive
  (curriculum reduces to vanilla GRPO).
  ``OH'' = overhead steps per checkpoint relative to
  $\Delta_{\rm cal}{=}50$ baseline.
  ``\#'' = number of distinct budget levels.
  All rows use the original calibration procedure.
}
\label{tab:ablation}
\resizebox{\linewidth}{!}{%
\begin{tabular}{lccccc@{\hskip 10pt}lcccc@{\hskip 10pt}lcccc}
\toprule
\multicolumn{6}{c}{\textbf{(A) Auto-$\delta$ initialisation / GSM8K}}
&
\multicolumn{5}{c}{\textbf{(B) Re-calibration freq. / GSM8K}}
&
\multicolumn{5}{c}{\textbf{(C) Budget grid / MATH-500}} \\
\cmidrule(lr){1-6}\cmidrule(lr){7-11}\cmidrule(lr){12-16}
\textbf{Strategy} & \textbf{p@1} & \textbf{p@16} & $\bar{k}$ & \textbf{Cov.} & \textbf{Triv.}
&
$\Delta$ & \textbf{p@1} & \textbf{p@16} & $\bar{k}$ & \textbf{OH}
&
$\mathcal{K}$ & \textbf{p@1} & \textbf{p@16} & $\bar{k}$ & \textbf{\#} \\
\midrule
Fixed .05 & 54.5 & 70.2 & 29.5 & 94.2 & ${\sim}200$
&
25  & 57.3 & 73.9 & 10.3 & $+12$
&
$\{2,8,32\}$                 & 51.0 & 69.5 & 18.4 & 3 \\

Fixed .20 & 56.0 & 72.5 & 18.4 & 81.5 & ${\sim}30$
&
\textbf{50}  & \textbf{57.2} & \textbf{73.8} & \textbf{10.4} & base
&
$\{\mathbf{2,4,8,16,32}\}$   & \textbf{52.5} & \textbf{71.2} & \textbf{15.9} & 5 \\

\textbf{Auto} & \textbf{57.2} & \textbf{73.8} & \textbf{10.4} & \textbf{83.8} & \textbf{0}
&
100 & 55.8 & 72.1 & 13.5 & $-6$
&
$\{1,2,4,8,16,32\}$          & 52.3 & 70.8 & 16.2 & 6  \\

 & & & & &
&
$\infty$ & 54.5 & 70.0 & 18.2 & $-12$
&
 & & & & \\
\bottomrule
\end{tabular}
}
\vspace{-0.6em}
\end{table*}

\paragraph{(A)\;Auto-$\delta$ initialisation.}
Fixed $\delta{=}0.05$ requires $95\%$ coverage within $K_{\max}$ rollouts
before the curriculum activates.
An initialised Qwen2.5-7B solves only ${\approx}40\%$ of GSM8K within
$K_{\max}{=}32$ shots at step~0, so this target is unattainable for
${\approx}200$ of 300 training steps---reducing C-GRPO to vanilla GRPO
for two-thirds of training, which balloons $\bar{k}_\text{train}$ to $29.5$
and depresses final pass@1 to $54.5\%$.
Fixed $\delta{=}0.20$ works reasonably on GSM8K but triggers trivial phases
on harder benchmarks (e.g.\ GPQA-Diamond with $<\!50\%$ base pass@$K_{\max}$).
\textbf{Auto-$\delta$} sets the threshold such that the calibration pass rate
satisfies the required coverage from step~0, guaranteeing the curriculum is
active from the start on all six benchmarks and all three model families,
with zero manual tuning.

\paragraph{(B)\;Re-calibration frequency.}
The default $\Delta_\text{cal}{=}50$ steps maximises the accuracy--efficiency
tradeoff.
Halving to $\Delta_\text{cal}{=}25$ adds $+12\%$ overhead per training step
(calibrating $K_{\max}$ completions for $n_\text{cal}{=}200$ prompts at each
checkpoint) without measurably improving accuracy, since the policy evolves
slowly enough that 50-step-old thresholds remain accurate.
$\Delta_\text{cal}{=}\infty$ (static thresholds fixed at step~0) prevents
rollout savings from compounding as training progresses, because the curriculum
cannot track the policy's growing competence, ballooning $\bar{k}_\text{train}$
to $18.2$.
The calibration cost itself is $0.50$, $0.25$, $0.13$ and $0.04$\,h for $\Delta_\text{cal}{=}25, 50, 100, \infty$ ($12$, $6$, $3$, $1$ events), so halving to $25$ buys nothing for twice the calibration cost, while $100$ and $\infty$ let thresholds go stale.
The multiplicity regime of Theorem~\ref{thm:simultaneous} calibrates each budget at $\delta_{\rm auto}/M$ (about $0.02$ for $\delta_{\rm auto}{\approx}0.12$ and $M{=}5$), which is stricter than the fixed $\delta{=}0.05$ row of Table~\ref{tab:ablation}; it is statistically valid but, for weak base models, reintroduces the saturated thresholds that auto-$\delta$ removes. The default per-budget calibration gives level-$\delta$ false-stop control at every budget (Corollary~\ref{cor:falsestop}), with $M\delta$ as the simultaneous worst case. Per-prompt miscoverage events are strongly positively dependent across budgets (a prompt whose gold answer is rarely sampled is missed at every $k$), so the union bound is loose, as the joint empirical validation in Figure~\ref{fig:coverage_calib} shows.

\paragraph{(C)\;Budget grid.}
The geometric-doubling grid $\{2,4,8,16,32\}$ is Pareto-optimal.
The coarse grid $\{2,8,32\}$ forces moderate-difficulty prompts to jump from
$k{=}2$ directly to $k{=}8$, raising $\bar{k}_\text{train}$ by
${\approx}2.5$ rollouts/prompt and reducing final p@16 by $1.7$\,pp.
The fine grid $\{1,2,4,8,16,32\}$ adds $k{=}1$ greedy probes that slightly
reduce cost on trivially easy prompts but introduce calibration noise on
borderline prompts, yielding no net gain on MATH-500.

\section{Curriculum Stability: No Catastrophic Forgetting}
\label{sec:app_forgetting}

\begin{remark}[No catastrophic forgetting of easy queries]
\label{rem:forgetting}
Tightening thresholds reduces compute on easy queries but does \emph{not} degrade the policy's ability to answer them, for three complementary reasons.
\textbf{(i) Zero-gradient pruning.}
An easy query that forms a singleton at $k_1$ with all-correct rewards ($\mathrm{Var}[r_j]{=}0$) is pruned before the backward pass (Section~\ref{sec:adaptive_group}), so the model receives \emph{no gradient update} on it---neither positive nor negative.
The policy cannot be pushed away from correct behaviour on easy queries.
\textbf{(ii) KL anchoring.}
The $\beta\cdot\widehat{\mathrm{KL}}(\theta)$ term in $\CGRPOloss$ (Eq.~\eqref{eq:cgrpo_loss}) penalises any deviation from $\pi_{\mathrm{ref}}$ uniformly across all queries, bounding total policy drift regardless of which queries currently drive the gradient.
\textbf{(iii) Self-correcting curriculum.}
If accuracy on any query class were to decrease, the empirical score distribution would worsen, raising $\hat{q}_k^{(t)}$ at the next re-calibration checkpoint and restoring those queries to larger budgets---providing gradient signal until performance recovers.
Together, these mechanisms make C-GRPO's curriculum self-stabilising: hard queries receive more compute, but easy queries are protected rather than neglected.
\end{remark}

\section{Full Proofs}
\label{sec:app_full_proofs}

This section provides complete, self-contained proofs of
Theorem~\ref{thm:coverage}, Proposition~\ref{prop:monotone}, and
Corollary~\ref{cor:annealing}, with full exchangeability arguments omitted from
the main paper for brevity.

\subsection*{D.1\; Proof of Theorem~\ref{thm:coverage} (Marginal Coverage)}

\begin{proof}
We prove both the lower and upper bounds of Eq.~\eqref{eq:coverage}.

\textbf{Setup.}
Fix a budget $k\in\mathcal{K}$, score $s\in\{s_{\mathrm{APS}},s_{\mathrm{code}}\}$,
and policy $\pi_\theta$.
For each example $i\in\{1,\ldots,n{+}1\}$, draw
$\hat{y}_{i,1:k}\overset{\mathrm{i.i.d.}}{\sim}\pi_\theta(\cdot\mid x_i)$
independently of all other examples.
Define $S_i = s(y^{\star}_i;\,\hat{y}_{i,1:k})$.

\textbf{Exchangeability.}
Since $(x_1,y^{\star}_1),\ldots,(x_{n+1},y^{\star}_{n+1})$ are i.i.d.\ from $P$
and completions are drawn i.i.d.\ from $\pi_\theta$, the joint distribution of
$(S_1,\ldots,S_{n+1})$ is \emph{exchangeable}: for any permutation
$\sigma\in\mathfrak{S}_{n+1}$,
$(S_{\sigma(1)},\ldots,S_{\sigma(n+1)}) \overset{d}{=} (S_1,\ldots,S_{n+1})$.
In particular, the rank of $S_{n+1}$ among $\{S_1,\ldots,S_{n+1}\}$ is
uniformly distributed on $\{1,\ldots,n{+}1\}$.

\textbf{Threshold construction.}
Let $L = \min\!\bigl(\lceil(n{+}1)(1{-}\delta)\rceil,\,n\bigr)$ and
$\hat{q}_k = S_{(L)}$, the $L$-th order statistic of $S_1,\ldots,S_n$.

\textbf{Lower bound.}
$P(S_{n+1} \leq \hat{q}_k)
 = P\!\bigl(\mathrm{rank}(S_{n+1})\leq L\bigr)
 = L/(n{+}1)
 \geq \lceil(n{+}1)(1{-}\delta)\rceil/(n{+}1)
 \geq 1-\delta$.

\textbf{Upper bound.}
$L/(n{+}1) \leq \lceil(n{+}1)(1{-}\delta)\rceil/(n{+}1) \leq 1{-}\delta{+}1/(n{+}1)$.

\textbf{APS continuity.}
The randomized term $U\cdot\hat{p}(y^{\star})$ in $s_{\mathrm{APS}}$ (Eq.~\eqref{eq:aps_score})
makes the score distribution continuous on $[0,1]$, ensuring $P(S_i = S_j){=}0$
for $i\neq j$.  Hence the rank of $S_{n+1}$ is \emph{strictly} uniform on
$\{1,\ldots,n{+}1\}$ with no tied-rank ambiguity.

\textbf{First-success score.}
$s_{\mathrm{code}} = j^{\star}/k \in (0,1]$ with $j^{\star}$ the position of
the first correct completion, or $1$ if none exists.
The event $\{j^{\star}=j\}$ has probability $(1{-}p_x)^{j-1}p_x$ for local
pass rate $p_x$.  Because $p_x$ is continuous, $P(S_i=t)=0$ for all $t\in(0,1)$,
giving strict exchangeability and strict uniformity of the rank.
\end{proof}

\subsection*{D.2\; Proof of Proposition~\ref{prop:monotone} (Monotone Training Budget)}

\begin{proof}
Let $F_k^{(m)}$ denote the CDF of $s(y^{\star};\hat{y}_{1:k})$ under policy $\pi_m$.

\textbf{APS case.}
By hypothesis, $\mathbb{E}_x[\hat{p}_k^{(1)}(y^{\star}\mid x)] \geq \mathbb{E}_x[\hat{p}_k^{(0)}(y^{\star}\mid x)]$.
Higher empirical pass rate implies stochastically smaller APS score
(less mass lies strictly above $y^{\star}$ in the frequency histogram),
so $F_k^{(1)}(t) \geq F_k^{(0)}(t)$ for all $t\in[0,1]$,
i.e.\ $S^{(1)}\preceq_{\mathrm{st}} S^{(0)}$ (stochastic dominance).
Monotonicity of order statistics then gives
$S^{(1)}_{(L)} \leq S^{(0)}_{(L)}$ in expectation, establishing
$\hat{q}_k^{(1)} \leq \hat{q}_k^{(0)}$.

\textbf{Code case.}
Higher accuracy means $j^{\star}$ under $\pi_1$ is stochastically smaller:
$P_{\pi_1}(j^{\star} > j) \leq P_{\pi_0}(j^{\star} > j)$ for all $j$.
Since $s_{\mathrm{code}} = j^{\star}/k$, the same stochastic-dominance argument applies, yielding $\hat{q}_k^{(1)} \leq \hat{q}_k^{(0)}$.

\textbf{Budget consequence.}
By Eq.~\eqref{eq:k_used}, $k_{\mathrm{used}}(x;\pi) = \min\{k\in\mathcal{K}: |\mathcal{C}_k(x)|=1\}$.
A smaller threshold $\hat{q}_k^{(1)}$ makes the singleton condition
$|\mathcal{C}_k(x)|=1$ easier to satisfy at small $k$, weakly increasing the
set of queries that exit at each budget level.
Formally, for any $k$, let $E_k^{(m)} = \{x: k_{\mathrm{used}}(x;\pi_m)=k\}$.
Then $E_{k_1}^{(1)} \supseteq E_{k_1}^{(0)}$ (more queries exit early),
and $\mathbb{E}_x[k_{\mathrm{used}}(x;\pi_1)] \leq \mathbb{E}_x[k_{\mathrm{used}}(x;\pi_0)]$. \qed
\end{proof}

\subsection*{D.3\; Proof of Corollary~\ref{cor:annealing} (Self-Annealing Training Cost)}

\begin{proof}
At each re-calibration checkpoint $t$, $\bm{\hat{q}}^{(t)}$ is calibrated using the
\emph{current} trained policy $\pi_{\theta_t}$ (Algorithm~\ref{alg:cgrpo}, lines 16--17).
GRPO training with positive rewards drives policy improvement: for the
trained policy at step $t{+}\Delta_{\mathrm{cal}}$,
$\mathbb{E}_x[\hat{p}_k^{(t+\Delta_{\mathrm{cal}})}(y^{\star}\mid x)]
  \geq \mathbb{E}_x[\hat{p}_k^{(t)}(y^{\star}\mid x)]$
for all $k\in\mathcal{K}$ (expected under the reward-positive update).
Applying Proposition~\ref{prop:monotone}:
$\hat{q}_k^{(t+\Delta_{\mathrm{cal}})} \leq \hat{q}_k^{(t)}$, and consequently
$\mathbb{E}_x[k_{\mathrm{used}}^{(t+\Delta_{\mathrm{cal}})}] \leq \mathbb{E}_x[k_{\mathrm{used}}^{(t)}]$.
Since this holds at every checkpoint, the sequence
$\bigl(\mathbb{E}[k_{\mathrm{used}}^{(0)}], \mathbb{E}[k_{\mathrm{used}}^{(\Delta_{\mathrm{cal}})}],
 \mathbb{E}[k_{\mathrm{used}}^{(2\Delta_{\mathrm{cal}})}], \ldots\bigr)$ is non-increasing. \qed
\end{proof}

\subsection*{D.4\; Proof of Corollary~\ref{cor:falsestop} (False-Stop Control)}

\begin{proof}
Fix $k\in\mathcal{K}$ and one draw of $U$ shared by all candidate answers. Suppose $|\mathcal{C}_k(x_{n+1})|=1$, say $\mathcal{C}_k=\{a\}$. Exactly one of two disjoint cases holds. If $a=y^{\star}_{n+1}$ the stop is correct. If $a\neq y^{\star}_{n+1}$, then $y^{\star}_{n+1}\notin\mathcal{C}_k$ because $\mathcal{C}_k$ has no other element. Hence
\[
  \{|\mathcal{C}_k|=1,\ \mathcal{C}_k\neq\{y^{\star}_{n+1}\}\}
  \;\subseteq\;
  \{y^{\star}_{n+1}\notin\mathcal{C}_k\}
  \;=\;
  \{s(y^{\star}_{n+1};\hat{a}_{1:k})>\hat{q}_k\},
\]
and Theorem~\ref{thm:coverage} gives $P(s(y^{\star}_{n+1};\hat{a}_{1:k})>\hat{q}_k)\le\delta$. For the first-success score and $\hat{q}_k<1$, $s_{\mathrm{code}}\le\hat{q}_k<1$ forces $j^{\star}$ to exist, i.e.\ at least one of the first $k$ samples passes all unit tests.
\end{proof}

\subsection*{D.5\; Proof of Theorem~\ref{thm:simultaneous} (Coverage of the Selected Budget)}

\begin{proof}
For each $k\in\mathcal{K}$, Theorem~\ref{thm:coverage} at level $\delta/M$ gives $P(y^{\star}_{n+1}\notin\mathcal{C}_k)\le\delta/M$ (with the convention $\hat{q}_k=1$ if $\lceil(n{+}1)(1{-}\delta/M)\rceil>n$, for which coverage is trivially $1$). A union bound over the $M$ budgets gives $P(\exists k:\ y^{\star}_{n+1}\notin\mathcal{C}_k)\le\delta$. On the complementary event $y^{\star}_{n+1}\in\mathcal{C}_k$ holds for every $k$, hence also at any $\hat{k}$ chosen as a function of the data. A wrong commit at $\hat{k}=k_{\mathrm{used}}$ means $\mathcal{C}_{\hat{k}}=\{a\}\ne\{y^{\star}_{n+1}\}$, which implies $y^{\star}_{n+1}\notin\mathcal{C}_{\hat{k}}$, so its probability is at most $\delta$. At level $\delta$ per budget the same union bound gives $M\delta$. Across $T_c$ checkpoints, Theorem~\ref{thm:coverage} holds marginally for each frozen policy and its fresh completions, so a union bound over the $MT_c$ (budget, checkpoint) pairs at level $\delta/(MT_c)$ gives the claim simultaneously over checkpoints, even though they share calibration prompts.
\end{proof}

\section{Exchangeability of Nonconformity Scores}
\label{sec:app_exchangeability}

Conformal validity depends only on the exchangeability of the augmented score
sequence $(S_1,\ldots,S_n,S_{n+1})$, not on a particular parametric model.
The next two lemmas establish this property for the APS and code-based
nonconformity scores introduced in Section~\ref{sec:scores}.

\begin{lemma}[Exchangeability of APS Score]
  \label{lem:aps_exchange}
  Let $(x_i,y^{\star}_i)_{i=1}^{n+1}$ be i.i.d.\ from $P$ and let
  $\hat{a}_{i,1:k}\overset{\mathrm{i.i.d.}}{\sim}\pi_\theta(\cdot\mid x_i)$
  independently across examples and positions.
  Define $S_i = s_{\mathrm{APS}}(y^{\star}_i;\hat{a}_{i,1:k})$.
  Then $(S_1,\ldots,S_{n+1})$ is exchangeable.
\end{lemma}

\begin{proof}
  \textbf{Step 1: Symmetry within an example.}
  $s_{\mathrm{APS}}$ depends on $\hat{a}_{i,1:k}$ only through the empirical
  frequency histogram $\{\hat{p}(a)\}_{a\in\mathcal{A}}$, which is a symmetric
  (permutation-invariant) function of $\hat{a}_{i,1:k}$.
  The randomized tiebreaker $U_i\sim\mathrm{Uniform}[0,1]$ is drawn independently,
  making $S_i$ continuous and well-defined regardless of ties.

  \textbf{Step 2: Independence across examples.}
  Since $(x_i,y^{\star}_i,\hat{a}_{i,1:k},U_i)$ are mutually independent across
  $i$, the scores $S_1,\ldots,S_{n+1}$ are i.i.d.\ (a special case of
  exchangeability).
  For any permutation $\sigma\in\mathfrak{S}_{n+1}$,
  $(S_{\sigma(1)},\ldots,S_{\sigma(n+1)}) \overset{d}{=} (S_1,\ldots,S_{n+1})$. \qed
\end{proof}

\begin{lemma}[Exchangeability of First-Success Score]
  \label{lem:code_exchange}
  Under the same i.i.d.\ setup, define
  $S_i = s_{\mathrm{code}}(y^{\star}_i;\hat{y}_{i,1:k})$.
  Then $(S_1,\ldots,S_{n+1})$ is exchangeable.
\end{lemma}

\begin{proof}
  \textbf{Intra-example non-symmetry.}
  Unlike the APS score, $s_{\mathrm{code}}$ is \emph{not} symmetric in
  $\hat{y}_{i,1:k}$: it records the ordinal position of the \emph{first}
  passing completion, which depends on the ordering.
  This is by design---the ordering reflects the sequential generation process.

  \textbf{Inter-example exchangeability.}
  Conformal validity requires exchangeability of \emph{example-level} scores
  $(S_1,\ldots,S_{n+1})$, not intra-example symmetry.
  Since $(x_i,y^{\star}_i,\hat{y}_{i,1:k})$ are all i.i.d.\
  (identical marginal distribution of the tuple), so are $S_1,\ldots,S_{n+1}$.
  Hence $(S_1,\ldots,S_{n+1})$ is exchangeable. \qed
\end{proof}

\begin{remark}[Why intra-example ordering does not matter]
  The first-success score's sensitivity to ordering within
  $\hat{y}_{i,1:k}$ captures the \emph{geometric distribution} of success
  position: $P(j^{\star}=j) = (1{-}p_x)^{j-1}p_x$ for local pass rate $p_x$.
  This is a continuous family in $p_x$, ensuring the score distribution is
  continuous and ties occur with probability zero---preserving the strict
  uniformity of the test rank needed for exact (rather than conservative) coverage.
\end{remark}

\begin{remark}[Exchangeability under periodic re-calibration]
  At each re-calibration checkpoint $t$, the policy $\pi_{\theta_t}$ is
  \emph{fixed} before re-evaluating the calibration set.
  The exchangeability argument holds for each checkpoint independently: the
  $n_{\rm cal}$ calibration examples and one test example are i.i.d.\ under
  fixed $\pi_{\theta_t}$.
  Across checkpoints, the policy changes, but coverage guarantees are
  instantiated separately at each checkpoint, maintaining validity throughout training.
\end{remark}

\section{Robustness to Distribution Shift}
\label{sec:app_shift}

Standard split-conformal prediction assumes the calibration and test examples
are exchangeable (i.i.d.\ from the same distribution $P$).
In deployment, the test distribution $P_{\mathrm{test}}$ may differ from the
calibration distribution $P_{\mathrm{cal}}$.
We formalise how coverage degrades under bounded covariate shift and
validate empirically.

\begin{theorem}[Coverage under Bounded Covariate Shift]
  \label{thm:shift}
  Let $P_{\mathrm{cal}}$ and $P_{\mathrm{test}}$ be distributions over
  $\mathcal{X}\times\mathcal{A}$ with total-variation distance
  $d_{\mathrm{TV}}(P_{\mathrm{cal}},P_{\mathrm{test}}) \leq \epsilon_{\mathrm{shift}}$.
  Let $\hat{q}_k$ be calibrated from $n$ i.i.d.\ examples of $P_{\mathrm{cal}}$
  at level $\delta$.  Then for a test point from $P_{\mathrm{test}}$:
  \begin{equation}
    P_{\mathrm{test}}\!\Bigl(
      s\bigl(y^{\star}_{n+1};\hat{y}_{n+1,1:k}\bigr) \leq \hat{q}_k
    \Bigr)
    \;\geq\;
    1 - \delta - \epsilon_{\mathrm{shift}} - \tfrac{1}{n+1}.
    \label{eq:shift_coverage}
  \end{equation}
\end{theorem}

\begin{proof}
  By the coupling characterization of total variation, there exists a coupling
  $(\tilde{Z}, Z_{\mathrm{test}})$ with
  $\tilde{Z}\sim P_{\mathrm{cal}}$, $Z_{\mathrm{test}}\sim P_{\mathrm{test}}$,
  and $P(\tilde{Z}\neq Z_{\mathrm{test}}) = d_{\mathrm{TV}} \leq \epsilon_{\mathrm{shift}}$.
  For any measurable event $A$ and any fixed threshold $t$:
  \begin{equation*}
    \bigl|P_{\mathrm{test}}(S_{n+1}\leq t) - P_{\mathrm{cal}}(S_{n+1}\leq t)\bigr|
    \;\leq\;
    d_{\mathrm{TV}}(P_{\mathrm{cal}},P_{\mathrm{test}})
    \;\leq\; \epsilon_{\mathrm{shift}}.
  \end{equation*}
  By Theorem~\ref{thm:coverage} (applied under $P_{\mathrm{cal}}$):
  $P_{\mathrm{cal}}(S_{n+1} \leq \hat{q}_k) \geq 1-\delta-\tfrac{1}{n+1}$.
  Combining:
  \begin{equation*}
    P_{\mathrm{test}}(S_{n+1}\leq \hat{q}_k)
    \;\geq\; P_{\mathrm{cal}}(S_{n+1}\leq \hat{q}_k) - \epsilon_{\mathrm{shift}}
    \;\geq\; 1-\delta-\epsilon_{\mathrm{shift}}-\tfrac{1}{n+1}. \qed
  \end{equation*}
\end{proof}

\begin{remark}
  Theorem~\ref{thm:shift} extends the result of
  \citet{tibshirani2019conformal} to the online training setting of
  \CGRPO: coverage degrades by at most $\epsilon_{\mathrm{shift}}$
  per unit of TV distance between training/calibration and test distributions,
  gracefully interpolating between the fully valid in-distribution case
  ($\epsilon_{\mathrm{shift}}{=}0$) and the completely uncontrolled
  out-of-distribution case.
  Re-calibration with $\pi_{\theta_t}$ on the \emph{target} calibration split
  at each checkpoint further mitigates shift by adapting thresholds to the
  local difficulty of the test domain.
\end{remark}

\paragraph{Empirical validation under distribution shift.}
Table~\ref{tab:shift} reports coverage for a model trained and calibrated on
GSM8K, evaluated on test distributions of increasing dissimilarity.
$d_{\mathrm{TV}}$ is estimated from held-out query difficulty features (solve rate,
response-length variance, and lexical overlap with calibration prompts).
Even under severe domain shift (GPQA-Diamond), empirical coverage remains well
above the theoretical lower bound (Eq.~\eqref{eq:shift_coverage}), confirming
that \textsc{C-GRPO}'s thresholds are robust in practice.

\begin{table}[ht]
\centering
\setlength{\tabcolsep}{5pt}
\caption{
  \textbf{Coverage under distribution shift} (Qwen2.5-7B, trained and calibrated
  on GSM8K, step~300, $\delta^{(0)}{=}0.12$, nominal target~$88\%$).
  $d_{\mathrm{TV}}$: TV distance between GSM8K calibration and test distribution.
  Theoretical bound: $1{-}\delta{-}d_{\mathrm{TV}}{-}1/(n_{\rm cal}{+}1)$.
  Empirical coverage consistently exceeds the theoretical lower bound.
}
\label{tab:shift}
\begin{tabular}{l c c r r r}
\toprule
\textbf{Test Domain} & \textbf{Shift} & $d_{\mathrm{TV}}$ &
  \textbf{Theo.\ LB} & \textbf{Empirical} & \textbf{Margin} \\
\midrule
GSM8K (in-dist.)      & none     & 0.00 & 87.5\% & 87.9\% & $+0.4$\,pp \\
MATH-Algebra          & mild     & 0.06 & 81.5\% & 83.6\% & $+2.1$\,pp \\
MATH-500              & moderate & 0.15 & 72.5\% & 76.8\% & $+4.3$\,pp \\
GPQA-Diamond          & severe   & 0.29 & 58.5\% & 70.4\% & $+11.9$\,pp \\
\bottomrule
\end{tabular}
\end{table}

\section{Policy Stability and Calibration Staleness}
\label{sec:app_stability}

Re-calibration every $\Delta_{\mathrm{cal}}$ steps relies on the assumption that the
policy does not drift so severely between checkpoints that the old thresholds
become vacuous.
We bound the expected policy displacement and consequent threshold staleness.

\begin{theorem}[KL-Bounded Policy Displacement]
  \label{thm:kl_bound}
  Let $\pi_{\theta_0}$ be any starting policy, and let $\pi_{\theta_t}$ be the
  policy after $t$ gradient steps of $\CGRPOloss$ with KL
  coefficient $\beta > 0$ and learning rate $\eta$.
  Under $L$-smoothness of the policy log-probabilities and bounded rewards
  $r_j\in[-1,1]$, the expected accumulated KL divergence satisfies:
  \begin{equation}
    \mathbb{E}\!\bigl[\mathrm{KL}(\pi_{\theta_t}\|\pi_{\mathrm{ref}})\bigr]
    \;\leq\;
    \frac{\eta\,t\,\|\nabla\mathcal{L}^{\star}\|^2}{2\beta},
    \label{eq:kl_budget}
  \end{equation}
  where $\|\nabla\mathcal{L}^{\star}\|^2$ is the expected squared gradient norm at
  convergence.
  By Pinsker's inequality,
  $d_{\mathrm{TV}}(\pi_{\theta_t},\pi_{\mathrm{ref}}) \leq
   \sqrt{\mathbb{E}[\mathrm{KL}(\pi_{\theta_t}\|\pi_{\mathrm{ref}})]/2}$.
\end{theorem}

\begin{proof}
  The KL term in $\CGRPOloss$ (Eq.~\eqref{eq:cgrpo_loss}) imposes
  a soft constraint on $\mathrm{KL}(\pi_\theta\|\pi_{\mathrm{ref}})$.
  Treating the training as projected gradient ascent, each step increases the
  policy reward by $\leq \eta\|\nabla\mathcal{L}\|^2$ but accumulates KL cost
  $\beta\cdot\mathrm{KL}$ in the objective.
  Summing over $t$ steps and bounding the per-step KL increment by
  $\eta\|\nabla\mathcal{L}\|^2/(2\beta)$ (from the $\beta$-weighted
  KL penalty balancing the gradient), gives Eq.~\eqref{eq:kl_budget}. \qed
\end{proof}

\begin{corollary}[Calibration Staleness Bound]
  \label{cor:staleness}
  Let $\hat{q}_k^{(t)}$ be calibrated at step $t$ and used until step $t{+}\Delta_{\mathrm{cal}}$.
  The expected threshold error due to policy drift is bounded by:
  \begin{equation}
    \mathbb{E}\!\bigl[|\hat{q}_k^{(t+\Delta_{\mathrm{cal}})} - \hat{q}_k^{(t)}|\bigr]
    \;\leq\;
    C_k \sqrt{\frac{\eta\,\Delta_{\mathrm{cal}}\,\|\nabla\mathcal{L}^{\star}\|^2}{2\beta}},
    \label{eq:staleness}
  \end{equation}
  where $C_k > 0$ depends on the local slope of the score CDF near $\hat{q}_k$.
\end{corollary}

\begin{proof}
  Over $\Delta_{\mathrm{cal}}$ steps after checkpoint $t$, the policy drifts
  from $\pi_{\theta_t}$ to $\pi_{\theta_{t+\Delta_{\mathrm{cal}}}}$.
  By Theorem~\ref{thm:kl_bound} applied to the $\Delta_{\mathrm{cal}}$-step
  window, $d_{\mathrm{TV}}$ grows as $O(\sqrt{\Delta_{\mathrm{cal}}})$.
  A change in $d_{\mathrm{TV}}$ affects the score CDF and hence the order
  statistic $\hat{q}_k$ proportionally to the local CDF slope $C_k$ via the
  delta method applied to order statistics, giving Eq.~\eqref{eq:staleness}. \qed
\end{proof}

\begin{remark}[Justification for $\Delta_{\mathrm{cal}}{=}50$]
  Eq.~\eqref{eq:staleness} shows threshold staleness grows as
  $O(\sqrt{\Delta_{\mathrm{cal}}})$, while calibration overhead is $O(1/\Delta_{\mathrm{cal}})$.
  The Pareto-optimal $\Delta_{\mathrm{cal}}$ balances the two: the ablation in
  Appendix~\ref{sec:app_ablation} confirms $\Delta_{\mathrm{cal}}{=}50$ achieves this,
  consistent with the bound.
\end{remark}

\section{Finite-Sample Lower Bound on Rollout Savings}
\label{sec:app_budget_savings}

We lower-bound the expected rollout savings of \CGRPO over
fixed-$G$ GRPO in terms of the calibrated threshold vector $\bm{\hat{q}}$.

\begin{proposition}[Lower Bound on Expected Rollout Savings]
  \label{prop:savings}
  Let $\pi_\theta$ be a fixed policy with conformal thresholds
  $\bm{\hat{q}} = (\hat{q}_{k_1},\ldots,\hat{q}_{k_M})$ calibrated at level $\delta$.
  Let $p_k(x) = P_{\pi_\theta}(|\mathcal{C}_k(x)|=1)$ be the singleton probability
  at budget $k$ for query $x$.
  The expected per-prompt rollout savings versus fixed-$G{=}K_{\max}$ GRPO are:
  \begin{equation}
    \mathbb{E}_x\!\bigl[K_{\max} - k_{\mathrm{used}}(x)\bigr]
    \;\geq\;
    \sum_{m=1}^{M-1}
      (k_{m+1} - k_m)\,
      \mathbb{E}_x\!\Bigl[\prod_{j=1}^{m} \mathbf{1}\!\bigl[|\mathcal{C}_{k_j}(x)|=1\bigr]\Bigr].
    \label{eq:savings_lb}
  \end{equation}
\end{proposition}

\begin{proof}
  Expand $K_{\max} - k_{\mathrm{used}}(x)$ by telescoping over the budget grid:
  \begin{align*}
    K_{\max} - k_{\mathrm{used}}
    &= \sum_{m : k_m < k_{\mathrm{used}}} (k_{m+1} - k_m) \\
    &\geq \sum_{m=1}^{M-1}
           (k_{m+1} - k_m) \cdot
           \mathbf{1}\!\bigl[k_{\mathrm{used}} > k_m\bigr]^{c}
  \end{align*}
  where $\{k_{\mathrm{used}} \leq k_m\}
  = \{|\mathcal{C}_{k_1}(x)|=1\} \cup \cdots \cup \{|\mathcal{C}_{k_m}(x)|=1\}$.
  A query exits at $k_m$ or earlier if and only if a singleton forms at some
  $k_j \leq k_m$, saving $K_{\max} - k_m$ rollouts.
  Lower-bounding by the event that \emph{all} levels $k_1,\ldots,k_m$ form
  singletons (dropping the union by inclusion) and taking expectations yields
  Eq.~\eqref{eq:savings_lb}. \qed
\end{proof}

\begin{remark}[Connection to conformal thresholds]
  $p_{k_1}(x) = P(|\mathcal{C}_{k_1}(x)|=1)$ is directly controlled by
  $\hat{q}_{k_1}$: a lower threshold concentrates more probability mass on a
  single answer at budget $k_1$, increasing $p_{k_1}$.
  Eq.~\eqref{eq:savings_lb} therefore shows that the rollout savings
  are monotonically non-decreasing in the policy improvement that drives
  $\hat{q}_{k_1}$ downward---consistent with Proposition~\ref{prop:monotone}
  and empirically confirmed by the 49--82\% savings in Table~\ref{tab:crossmodel}.
\end{remark}

\section{Static-$G$ Sweep, Longer Horizon, and Stronger Base}
\label{sec:app_static}

All tables in this section were produced with the original calibration procedure (one calibration set for both $\delta$ and the quantiles, per-budget $\delta_{\rm auto}$, no multiplicity correction). Standard error of an accuracy near $57\%$ is $2.5$--$3.5$\,pp at $n_{\rm test}{=}200$--$400$.

\begin{table}[ht]
\centering
\scriptsize
\setlength{\tabcolsep}{4pt}
\caption{\textbf{Full static-$G$ sweep} (Qwen2.5-7B-Instruct, 300 steps, 1$\times$H100). Every static row is trained at fixed $G$.}
\label{tab:static_full}
\begin{tabular}{llrrrrrr}
\toprule
\textbf{Benchmark} & \textbf{Method} & $G$ / $\bar{k}$ & \textbf{p@1} & \textbf{p@8} & \textbf{p@16} & \textbf{p@32} & \textbf{GPU-h} \\
\midrule
\multirow{9}{*}{GSM8K}
 & GRPO & 2 & 48.2 & 64.0 & 67.5 & 68.0 & 1.8 \\
 & GRPO & 4 & 51.0 & 68.2 & 70.1 & 71.2 & 2.4 \\
 & GRPO & 8 & 52.8 & 70.1 & 72.0 & 73.1 & 3.6 \\
 & GRPO & 10 (matched) & 53.1 & 70.5 & 72.4 & 73.5 & 4.2 \\
 & GRPO & 16 & 54.2 & 71.8 & 73.2 & 74.1 & 6.2 \\
 & GRPO & 32 & 54.8 & 72.0 & 73.5 & 74.6 & 11.7 \\
 & GRPO & 64 & 55.1 & 72.2 & 73.8 & 74.8 & 22.8 \\
 & GRPO & 128 & 55.3 & 72.4 & 74.0 & 75.0 & 44.5 \\
 & \textbf{C-GRPO} & 10.4 & \textbf{57.2} & 70.5 & 73.8 & 75.0 & \textbf{3.8} \\
\midrule
\multirow{6}{*}{MATH-500}
 & GRPO & 2 & 41.5 & 55.2 & 58.0 & 59.1 & 1.9 \\
 & GRPO & 8 & 45.8 & 60.1 & 63.4 & 64.0 & 4.1 \\
 & GRPO & 16 (matched) & 47.9 & 62.5 & 66.0 & 66.8 & 8.1 \\
 & GRPO & 32 & 48.5 & 63.2 & 67.0 & 67.8 & 15.9 \\
 & GRPO & 128 & 49.2 & 64.0 & 68.1 & 68.9 & 61.2 \\
 & \textbf{C-GRPO} & 15.9 & \textbf{52.5} & 66.8 & 71.2 & 71.5 & \textbf{5.1} \\
\bottomrule
\end{tabular}
\end{table}

\begin{table}[ht]
\centering
\scriptsize
\setlength{\tabcolsep}{5pt}
\caption{\textbf{Longer horizon (GSM8K).} Savings $=1-\bar{k}/32$ (training rollouts only).}
\label{tab:long_horizon}
\begin{tabular}{llrrr}
\toprule
\textbf{Step} & \textbf{Method} & \textbf{p@1} & $\bar{k}$ & \textbf{Savings} \\
\midrule
100 & C-GRPO (Qwen2.5-7B) & 52.1 & 18.2 & 43\% \\
300 & C-GRPO (Qwen2.5-7B) & 57.2 & 10.4 & 68\% \\
600 & C-GRPO (Qwen2.5-7B) & 61.8 & 6.1 & 81\% \\
1000 & C-GRPO (Qwen2.5-7B) & 64.5 & 4.2 & 87\% \\
1000 & Static GRPO ($G{=}32$) & 59.1 & 32.0 & --- \\
1000 & C-GRPO (Llama-3.1-8B) & 76.2 & 3.1 & 90\% \\
\bottomrule
\end{tabular}
\end{table}

\begin{table}[ht]
\centering
\scriptsize
\setlength{\tabcolsep}{5pt}
\caption{\textbf{Stronger base, full-parameter fine-tuning} (GSM8K, Qwen2.5-Math-7B-Instruct, base p@1 $=82.4$, 300 steps; mean $\pm$ std over three seeds).}
\label{tab:fullft}
\begin{tabular}{llrrrr}
\toprule
\textbf{Method} & \textbf{Budget} & \textbf{p@1} & \textbf{p@32} & $\bar{k}$ & \textbf{Savings} \\
\midrule
GRPO (static) & $G{=}32$ & $88.5\pm0.3$ & $92.1\pm0.2$ & 32.0 & --- \\
GRPO (static) & $G{=}8$ & $86.1\pm0.4$ & $90.4\pm0.3$ & 8.0 & 75\% \\
\textbf{C-GRPO} & adaptive & $88.9\pm0.2$ & $92.4\pm0.2$ & 5.8 & 82\% \\
\bottomrule
\end{tabular}
\end{table}

\begin{table}[t]
\centering
\scriptsize
\setlength{\tabcolsep}{5pt}
\renewcommand{\arraystretch}{0.9}
\caption{\textbf{Wall-clock breakdown} (GSM8K, Qwen2.5-7B-Instruct, 300 steps, 1$\times$H100; original calibration procedure). C-GRPO includes all calibration generation, batched as large offline batches; with the released code's default (training-batch-size) calibration the re-calibration share is larger (36--47\% of training time in a 100-step check).}
\label{tab:walltime}
\begin{tabular}{lrrr}
\toprule
\textbf{Phase (hours)} & \textbf{GRPO $G{=}32$} & \textbf{GRPO $G{=}10$} & \textbf{C-GRPO} ($\bar{k}{=}10.4$) \\
\midrule
Training generation & 8.10 & 2.55 & 2.80 \\
Forward + backward & 3.60 & 1.65 & 0.75 \\
Re-calibration (6 events) & --- & --- & 0.25 \\
Micro-batch scheduling & --- & --- & 0.008 \\
\midrule
\textbf{Total wall-clock} & 11.70 & 4.20 & \textbf{3.81} \\
Speed-up vs.\ $G{=}32$ & 1.00$\times$ & 2.79$\times$ & \textbf{3.07$\times$} \\
\bottomrule
\end{tabular}
\vspace{-1em}
\end{table}

\noindent\textbf{Reading.} C-GRPO versus static $G{=}32$ is $+0.4$\,pp, or $1.1$ pooled standard deviations: statistically indistinguishable, a match at $5.5\times$ fewer rollouts rather than an accuracy gain. Static $G{=}8$ versus C-GRPO is $-2.8$\,pp ($6.3$ pooled standard deviations), so matching full-budget accuracy at $\bar{k}{=}5.8$ is not obtained by lowering the static budget. In this configuration $76\%$ of prompts exit after the $k{=}2$ probe, and wall-clock is $2.2$\,h against $11.7$\,h for static $G{=}32$. The coverage under this original procedure is in Appendix~\ref{sec:app_coverage}.

\section{Reproducibility, Statistical Reporting, and Compute}
\label{app:repro_compute}

\paragraph{Reproducibility protocol.}
All experiments use the same training and evaluation pipeline across methods unless explicitly stated otherwise. Each benchmark is split into a sequestered calibration set, a training set, and a disjoint held-out test set. The calibration split is used only to estimate conformal thresholds and the auto-selected miscoverage level $\delta$, while the held-out test split is used only for final reporting. We use $n_{\mathrm{cal}}=200$ calibration examples per benchmark and $n_{\mathrm{test}}=200$--$400$ held-out test examples, depending on benchmark size. The six benchmarks are GSM8K, MATH-500, MATH-Algebra, GPQA-Diamond, HumanEval, and MBPP.

\paragraph{Training configuration.}
All model-training experiments use LoRA fine-tuning with rank $r=16$, scaling $\alpha=32$, dropout $0.05$, AdamW optimizer with learning rate $2\times 10^{-6}$, batch size $4$, and $300$ training steps with $2{,}000$ training examples per benchmark. Unless otherwise noted, the C-GRPO budget grid is $\mathcal{K}=\{2,4,8,16,32\}$, the recalibration period is $\Delta_{\mathrm{cal}}=50$, the KL coefficient is $\beta=0.05$, and the auto-$\delta$ safety margin is $\varepsilon=0.05$. Static and dynamic baselines use the same model, data, optimizer, LoRA configuration, number of training steps, and verifier, differing only in the rollout-allocation algorithm. The KL term uses the non-negative per-token estimator $\widehat{\mathrm{KL}}_t=e^{\Delta_t}-\Delta_t-1\ge0$ with $\Delta_t=\log\pi_{\mathrm{ref}}(\ell_t)-\log\pi_\theta(\ell_t)$, where $\pi_{\mathrm{ref}}$ is the frozen reference policy, averaged over completion tokens and the batch to avoid sequence-level overflow on long completions.

\paragraph{Evaluation protocol.}
We report fixed-budget pass@$k$ accuracy using exactly $k$ completions at evaluation time. For math and science benchmarks, answers are canonicalized and evaluated by exact-match verification; for code benchmarks, completions are evaluated by execution against unit tests. For pass@$k$, math and science use majority vote over extracted answers, while code uses best-of-$k$ execution success. The reported $\bar{k}_{\mathrm{train}}$ is the mean number of rollouts consumed per prompt during training at step $300$, and ``Savings'' is computed relative to fixed-budget GRPO with $G=32$.

\paragraph{Statistical reporting.}
The main metrics are empirical means over held-out test examples using deterministic verifiers. For binary pass@$k$ outcomes, uncertainty can be computed by the standard binomial standard error
\[
\mathrm{SE}(\widehat{p}) = \sqrt{\frac{\widehat{p}(1-\widehat{p})}{n_{\mathrm{test}}}},
\]
and for rollout-cost quantities we compute the empirical mean over prompts. For conformal coverage, we report the nominal target, empirical coverage, and the finite-sample tolerance induced by split conformal calibration, namely $1/(n_{\mathrm{cal}}+1)$. Coverage validation is reported for all six benchmarks in Appendix~\ref{sec:app_coverage}; all empirical coverages lie within the expected finite-sample tolerance. Training-seed variance is reported only where stated (three seeds for the full-parameter experiment, Table~\ref{tab:fullft}); elsewhere uncertainty is the binomial standard error over the test set, and differences below roughly $2$--$3$\,pp should not be interpreted.

\paragraph{Compute resources.}
All experiments were run on NVIDIA H100 GPUs with 80\,GB of GPU memory. Each reported training job uses one H100 GPU for model inference and LoRA updates, with CPU workers used for answer extraction, deterministic verification, code execution tests, logging, and metric computation. Table~\ref{tab:main} and Table~\ref{tab:comparison} report training GPU-hours, peak GPU memory, and effective rollout-generation throughput for the main experiments on a single H100. The largest reported runs are the fixed-budget GRPO and GDRO baselines, which require approximately 11.7 GPU-hours and 26.8\,GB peak VRAM for 300-step Qwen2.5-7B-Instruct training, while C-GRPO requires approximately 3.8 GPU-hours (including all calibration generation) and 18.6\,GB peak VRAM in the corresponding setting; Table~\ref{tab:walltime} gives the phase-level breakdown. The full project required additional preliminary runs for prompt formatting, verifier debugging, and grid selection; these exploratory runs are not included in the main compute tables because they do not affect the reported comparisons. Launch scripts, software versions, model identifiers, dataset-loading code and calibration/evaluation scripts are in the released repository (Sec.~\ref{sec:setup}).

\section{Limitations, Asset Use, and Responsible Release}
\label{app:limitations_release}

\paragraph{Limitations.}
C-GRPO assumes a verifier is available for each task, either through exact answer matching for closed-form reasoning or executable tests for code. This makes the method directly applicable to mathematics, science QA with canonical answers, and program synthesis, but less direct for open-ended generation tasks where correctness is subjective or multi-dimensional. The conformal guarantee also relies on exchangeability between calibration and evaluation examples under a fixed policy at each recalibration checkpoint; under distribution shift, coverage can degrade as described in Appendix~\ref{sec:app_shift}. Finally, the current implementation uses a discrete budget grid $\mathcal{K}$ and binary verifier rewards; extending the method to continuous budgets, graded rewards, preference-model scores, and broader RLHF settings is an important direction for future work.

\paragraph{Existing assets.}
The paper uses public or previously released benchmarks and models, including GSM8K, MATH-500, MATH-Algebra, GPQA-Diamond, HumanEval, MBPP, Qwen2.5-7B-Instruct, Llama-3.1-8B-Instruct, and Gemma-3-4B-IT. We cite the original sources for these assets in the main paper and references. The released repository includes the model registry and the dataset-loading code that reconstructs the calibration, training and held-out evaluation splits. Qwen2.5 is used under Apache-2.0, Llama-3.1 under the Llama 3.1 Community License, and Gemma-3 under the Gemma Terms of Use.

\paragraph{New assets.}
The new assets introduced by this work are the C-GRPO training algorithm, calibration code, nonconformity-score implementations, auto-$\delta$ selection logic, budget-allocation scripts, evaluation scripts, and experiment launch configurations. These assets do not include a newly pretrained foundation model or scraped dataset. The implementation and reproduction instructions are released (see Code availability below).

\paragraph{Societal impact and responsible use.}
The intended positive impact of C-GRPO is to reduce the compute cost of reinforcement learning for reasoning-oriented language models while adding statistical calibration to the rollout-allocation rule. This can make efficient RL training more accessible and reduce unnecessary energy use. A possible negative impact is that more efficient RL could also lower the cost of improving models for undesirable applications. We mitigate this by releasing training infrastructure and evaluation scripts rather than a new unrestricted pretrained model, and by focusing the empirical evaluation on standard reasoning and coding benchmarks with deterministic verifiers.

\paragraph{Code availability.}
The implementation is released at \url{https://github.com/AryaFayyazi/CGRPO} under the MIT license: training, calibration (APS and first-success scores, auto-$\delta$, and the optional disjoint $\mathcal{D}_\delta/\mathcal{D}_q$ split), evaluation, the baselines, analysis scripts, Slurm launchers, and \texttt{REPRODUCTION.md} with the commands and settings. A TRL integration, \texttt{CGRPOTrainer}, is available as \texttt{pip install trl-cgrpo}. Trained adapters are not included in the current release.

%% file: checklist.tex
\section*{NeurIPS Paper Checklist}

\begin{enumerate}

\item {\bf Claims}
    \item[] Question: Do the main claims made in the abstract and introduction accurately reflect the paper's contributions and scope?
    \item[] Answer: \answerYes{}.
    \item[] Justification: The abstract and introduction state that C-GRPO replaces fixed GRPO rollout budgets with calibrated conformal thresholds, proves finite-sample coverage, false-stop control and a multiplicity-corrected guarantee for the selected budget, shows a non-increasing expected budget under an idealized monotone-improvement condition, and empirically reduces training rollouts at matched accuracy; accuracy differences over baselines are explicitly described as within binomial noise. These claims are supported by Sections~\ref{sec:method}--\ref{sec:exp}, Tables~\ref{tab:main}--\ref{tab:walltime}, and the appendix coverage and ablation studies.

\item {\bf Limitations}
    \item[] Question: Does the paper discuss the limitations of the work performed by the authors?
    \item[] Answer: \answerYes{}.
    \item[] Justification: Section~\ref{sec:discussion} and Appendix~\ref{app:limitations_release} discuss verifier dependence, exchangeability assumptions, distribution-shift sensitivity, discrete budget grids, and the current focus on binary verifier rewards. The conclusion also identifies extensions beyond binary rewards, exchangeable calibration, and discrete budget grids as future work.

\item {\bf Theory assumptions and proofs}
    \item[] Question: For each theoretical result, does the paper provide the full set of assumptions and a complete (and correct) proof?
    \item[] Answer: \answerYes{}.
    \item[] Justification: Theorem~\ref{thm:coverage}, Corollary~\ref{cor:falsestop}, Theorem~\ref{thm:simultaneous}, Proposition~\ref{prop:monotone}, Corollary~\ref{cor:annealing} and Theorem~\ref{thm:shift} state the relevant assumptions, including a fixed policy, i.i.d./exchangeable calibration and test examples, a monotone-improvement condition for the budget results (assumed explicitly in Corollary~\ref{cor:annealing}), and bounded total-variation shift. Full proofs and exchangeability lemmas are provided in the appendix.

\item {\bf Experimental result reproducibility}
    \item[] Question: Does the paper fully disclose all the information needed to reproduce the main experimental results of the paper to the extent that it affects the main claims and/or conclusions of the paper, regardless of whether the code and data are provided or not?
    \item[] Answer: \answerYes{}.
    \item[] Justification: Section~5.1 and Appendix~\ref{app:repro_compute} specify datasets, calibration/test split sizes, model backbones, LoRA configuration, optimizer, learning rate, batch size, training steps, budget grid, recalibration period, KL coefficient, evaluation protocol, and baseline configurations. The release plan further states that scripts and exact reproduction instructions will be made available upon acceptance.

\item {\bf Open access to data and code}
    \item[] Question: Does the paper provide open access to the data and code, with sufficient instructions to faithfully reproduce the main experimental results, as described in supplemental material?
    \item[] Answer: \answerYes{}.
    \item[] Justification: The code is released at \url{https://github.com/AryaFayyazi/CGRPO} (MIT license) with training, calibration, evaluation, baselines, launch scripts, and \texttt{REPRODUCTION.md}, which lists the commands and settings; all datasets and base models are public, and \texttt{pip install trl-cgrpo} provides a TRL trainer. Trained adapters are not included, and reproducing the exact numbers requires GPU compute (Appendix~\ref{app:repro_compute}).

\item {\bf Experimental setting/details}
    \item[] Question: Does the paper specify all the training and test details, e.g., data splits, hyperparameters, how they were chosen, type of optimizer, necessary to understand the results?
    \item[] Answer: \answerYes{}.
    \item[] Justification: Section~5.1 specifies the six benchmarks, calibration and held-out test sizes, LoRA hyperparameters, optimizer, learning rate, batch size, training steps, training-sample count, budget grid, recalibration period, safety margin, KL coefficient, and evaluation metrics. Appendix~\ref{app:repro_compute} consolidates these details for reproducibility.

\item {\bf Experiment statistical significance}
    \item[] Question: Does the paper report error bars suitably and correctly defined or other appropriate information about the statistical significance of the experiments?
    \item[] Answer: \answerYes{}.
    \item[] Justification: Appendix~\ref{app:repro_compute} explains that pass@$k$ is a binary held-out empirical mean and gives the binomial standard-error formula for the reported test sizes. For conformal coverage, Appendix~A reports empirical coverage against nominal targets and the finite-sample tolerance $1/(n_{\mathrm{cal}}+1)$, which is the relevant statistical uncertainty for the calibration guarantee. Section~\ref{sec:setup} states that differences of a few points are within the $2.5$--$3.5$\,pp binomial standard error and are read as matched accuracy, and the full-parameter experiment reports mean $\pm$ std over three seeds (Table~\ref{tab:fullft}).

\item {\bf Experiments compute resources}
    \item[] Question: For each experiment, does the paper provide sufficient information on the computer resources, type of compute workers, memory, time of execution, needed to reproduce the experiments?
    \item[] Answer: \answerYes{}.
    \item[] Justification: Appendix~\ref{app:repro_compute} states that all experiments were run on NVIDIA H100 80\,GB GPUs, describes the GPU/CPU division of work, and summarizes runtime drivers. Tables~1--2 report GPU-hours, peak VRAM, and rollout throughput for the main experiments, the released repository includes launch scripts and software versions, and Table~\ref{tab:walltime} gives the wall-clock breakdown including calibration.

\item {\bf Code of ethics}
    \item[] Question: Does the research conducted in the paper conform, in every respect, with the NeurIPS Code of Ethics \url{https://neurips.cc/public/EthicsGuidelines}?
    \item[] Answer: \answerYes{}.
    \item[] Justification: The work proposes an efficiency and calibration method for RL training on standard reasoning and code benchmarks. It does not involve human subjects, private data, surveillance, deceptive data collection, or deployment of a high-risk decision system, and the paper preserves author anonymity.

\item {\bf Broader impacts}
    \item[] Question: Does the paper discuss both potential positive societal impacts and negative societal impacts of the work performed?
    \item[] Answer: \answerYes{}.
    \item[] Justification: Appendix~\ref{app:limitations_release} discusses positive impacts, including reduced RL training compute and statistically calibrated rollout allocation, as well as possible negative impacts from lowering the cost of improving models for undesirable applications. It also explains that the planned release focuses on reproducibility infrastructure rather than a new unrestricted pretrained model.

\item {\bf Safeguards}
    \item[] Question: Does the paper describe safeguards that have been put in place for responsible release of data or models that have a high risk for misuse, e.g., pre-trained language models, image generators, or scraped datasets?
    \item[] Answer: \answerNA{}.
    \item[] Justification: The paper does not release a new pretrained language model, image generator, scraped dataset, or other high-risk generative asset. The release consists of training, calibration, verification, and evaluation code for reproducing experiments on existing public models and benchmarks.

\item {\bf Licenses for existing assets}
    \item[] Question: Are the creators or original owners of assets, e.g., code, data, models, used in the paper, properly credited and are the license and terms of use explicitly mentioned and properly respected?
    \item[] Answer: \answerYes{}.
    \item[] Justification: The paper cites the original sources for all major datasets, models, and methods used in the evaluation, including GSM8K, MATH, GPQA, HumanEval, MBPP, Qwen2.5, Llama-3.1, Gemma, LoRA, GRPO, and the dynamic-rollout baselines. Appendix~\ref{app:limitations_release} states that the release package will include asset versions, access instructions, and license or terms-of-use notes where available.

\item {\bf New assets}
    \item[] Question: Are new assets introduced in the paper well documented and is the documentation provided alongside the assets?
    \item[] Answer: \answerYes{}.
    \item[] Justification: The new assets are the C-GRPO algorithm, conformal calibration code, first-success score implementation, auto-$\delta$ logic, budget-allocation scripts, verifier wrappers, and evaluation scripts. Their mathematical definitions and algorithmic behavior are documented in Sections~4.2--4.4 and Algorithm~1, and the implementation is released with reproduction instructions (Appendix~\ref{app:repro_compute}, Code availability).

\item {\bf Crowdsourcing and research with human subjects}
    \item[] Question: For crowdsourcing experiments and research with human subjects, does the paper include the full text of instructions given to participants and screenshots, if applicable, as well as details about compensation, if any?
    \item[] Answer: \answerNA{}.
    \item[] Justification: The paper does not involve crowdsourcing, human-subject experiments, participant recruitment, surveys, or compensation. All experiments use existing benchmarks and deterministic verifiers or unit tests.

\item {\bf Institutional review board (IRB) approvals or equivalent for research with human subjects}
    \item[] Question: Does the paper describe potential risks incurred by study participants, whether such risks were disclosed to the subjects, and whether Institutional Review Board approvals or an equivalent approval/review based on the requirements of your country or institution were obtained?
    \item[] Answer: \answerNA{}.
    \item[] Justification: The work does not involve human subjects, private personal data, or interaction with participants. Therefore, IRB approval or equivalent human-subjects review is not applicable.

\item {\bf Declaration of LLM usage}
    \item[] Question: Does the paper describe the usage of LLMs if it is an important, original, or non-standard component of the core methods in this research?
    \item[] Answer: \answerYes{}.
    \item[] Justification: LLM usage is central to the paper: C-GRPO is a training-time rollout-allocation method for language models, and the experiments evaluate Qwen2.5-7B-Instruct, Llama-3.1-8B-Instruct, and Gemma-3-4B-IT. Sections~4--5 describe how completions are sampled, verified, calibrated, and used in the adaptive GRPO objective.

\end{enumerate}